\documentclass[10pt,twocolumn,letterpaper]{article}

\usepackage{iccv}
\usepackage{times}
\usepackage{epsfig}
\usepackage{graphicx}
\usepackage{amsmath}
\usepackage{amssymb}
\usepackage{tabularx}
\usepackage{caption}
\usepackage{subcaption}
\usepackage{multirow}
\usepackage{makecell}
\usepackage{enumitem}
\usepackage{textcomp}

\usepackage[pagebackref=true,breaklinks=true,colorlinks,bookmarks=false]{hyperref}

\iccvfinalcopy

\def\mypar#1{\vspace{1mm}{\bf #1.}\hspace{1mm}}
\newcommand{\cmark}{\checkmark}

\usepackage{array}
\usepackage{calc}

\newcommand{\omitme}[1]{}

\newcommand{\cccr}[1]{} 

\ificcvfinal\fi
\begin{document}

\title{Unsupervised Image Decomposition in Vector Layers}

\author{
Othman Sbai$^{1,2}$
\and
Camille Couprie$^{1}$
\and
Mathieu Aubry$^{2}$
\and
$^{1}$ Facebook AI Research, $^{2}$ LIGM (UMR 8049) - \'Ecole des Ponts, UPE
}

\maketitle

\begin{abstract}
   Deep image generation is becoming a tool to enhance artists and designers creativity potential. In this paper, we aim at making the generation process more structured and easier to interact with. Inspired by vector graphics systems, we propose a new deep image reconstruction paradigm where the outputs are composed from simple layers, defined by their color and a vector transparency mask. This presents a number of advantages compared to the commonly used convolutional network architectures. In particular, our layered decomposition allows simple user interaction, for example to update a given mask, or change the color of a selected layer. From a compact code, our architecture also generates vector images with a virtually infinite resolution, the color at each point in an image being a parametric function of its coordinates. 
   We validate the efficiency of our approach by comparing reconstructions with state-of-the-art baselines given similar memory resources on CelebA and ImageNet datasets. Most importantly, we demonstrate several applications of our new image representation obtained in an unsupervised manner, including editing, vectorization and image search.
\end{abstract}

\section{Introduction}

\begin{figure}
    \centering
   \includegraphics[width=\linewidth]{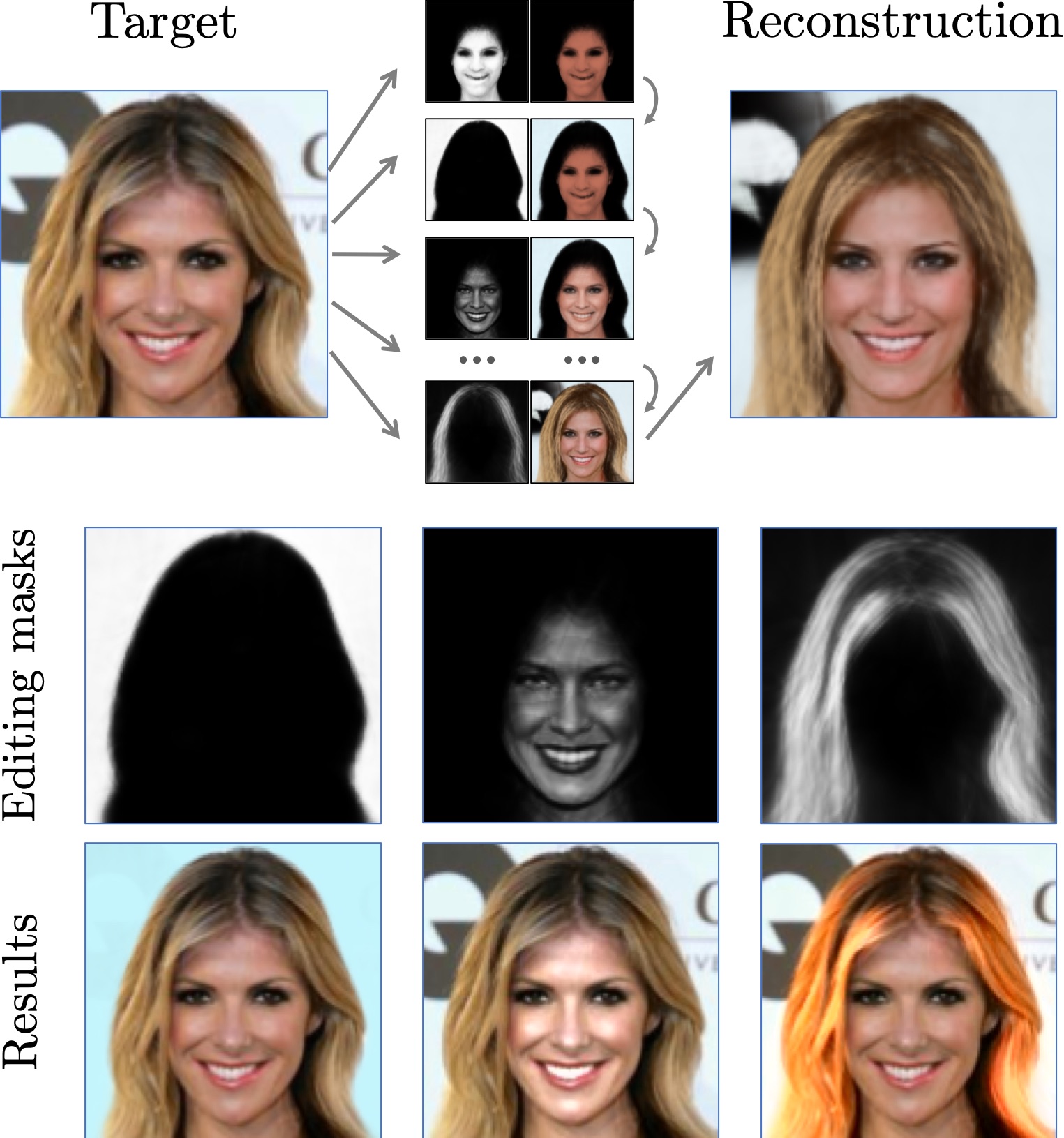}
    \caption{Our system learns in an unsupervised manner a decomposition of images as superimposed $\alpha$-channel masks (top) that can be used for quick image editing (bottom).}
    \label{fig:teaser}
\end{figure}

Deep image generation models demonstrate breathtaking and inspiring results, e.g. ~\cite{Zhu2017cycleGANs,Zhu2017bicycleGAN,karras2019style,biggan}, but usually offer limited control and little interpretability. It is indeed particularly challenging to learn end-to-end editable image decomposition without relying on either expensive user input or handcrafted image processing tools. In contrast, we introduce and explore a new deep image generation paradigm, which follows an approach similar to the one used in interactive design tools. We formulate image generation as the composition of successive layers, each associated to a single color.
Rather than learning high resolution image generation, we produce a decomposition of the image in vector layers, that can easily be used to edit images at any resolution.
We aim at enabling designers to easily build on the results of deep image generation methods, by editing layers individually, changing their characteristics, or intervening in the middle of the generation process.

Our approach is in line with the long standing Computer Vision trend to look for a simplified and compact representation of the visual world. For examples, in 1971 Binford~\cite{binford1971visual} proposes to represent the 3D world using generalized cylinders and in 1987 the seminal work of Biederman~\cite{biederman1987recognition} aims at explaining and interpreting the 3D world and images using geons, a family of simple parametric shapes. These ideas have recently been revisited using Neural Networks to represent a 3D shape using a set of blocks~\cite{tulsiani2017learning} or, more related to our approach, a set of parametric patches~\cite{groueix2018papier}.
The idea of identifying elementary shapes and organizing them in layers has been successfully applied to model images~\cite{adelson1991layered,SceneCollaging} and videos~\cite{wang1994representing}. 
A classical texture generation method, the dead leaves model~\cite{lee2001occlusion} which creates realistic textures by relying on the iteration of simple patterns addition, is particularly related to our work.

We build on this idea of composing layers of simple primitives in order to design a deep image generation method, relying on two core ingredients. 
First, the learning of vector transparency masks as parametric continuous function defined on the unit square. In practice, this function is computed by a network applied at 2D coordinates on a square grid, to output mask values at each pixel coordinates.
Second, a mask blending module which we use to iteratively build the images by superimposing a mask with a given color to the previous generation.
At each step of our generation process, a network predicts both parameters and color for one mask. Our final generated image is the result of blending a fixed number of colored masks. One of the advantages of this approach is that, differently to most existing deep generation setups where the generation is of fixed size, our generations are vector images defined continuously, and thus have virtually infinite resolution. Another key aspect is that the generation process is easily interpretable, allowing simple user interaction.

To summarize, our main contribution is a new deep image generation paradigm which:
\begin{itemize}
    \item builds images iteratively from masks corresponding to meaningful image regions, learned without any semantic supervision. 
    \item is one of the first to generate {\it vector} images from a compact code. 
    \item is useful for several applications, including image editing using generated masks, image vectorization, and image search in mask space.
\end{itemize}
Our code will be made available\footnote{http://imagine.enpc.fr/\texttildelow sbaio/pix2vec/}.

\section{Related work}

We begin this section by presenting relevant works on image vectorization, then focus on most related unsupervised image generation strategies and finally discuss applications of deep learning to image manipulation.

\mypar{Vectorization}
Many vector-based primitives have been proposed to allow shape editing, color editing and vector image processing ranging from paths and polygons filled with uniform color or linear and radial gradients~\cite{richardt2014vectorising,favreau2017photo2clipart}, to region based partitioning using triangulation~\cite{demaret2006image,liao2012subdivision,duan2015image}, parametric patches (Bezier patches)~\cite{xia2009patch} or diffusion curves~\cite{orzan2008diffusion}. 
We note that traditionally, image vectorization techniques were handcrafted using image smoothing and edge detectors. 
In contrast, our approach parametrizes the image using a function defined by a neural network. 

Differentiable image parametrizations with neural networks were first proposed by Stanley et al.~\cite{stanley2007compositional} which introduced Compositional Pattern Producing Networks (CPPNs) that are simply neural networks that map \omitme{(x,y)} pixel coordinates to \omitme{(r,g,b)} image colors at each pixel. The architecture of the network determines the space of images that can be generated. Since CPPNs learn images as functions of pixel coordinates they provide the ability to sample images at high resolution. The weights of the network can be optimized to reconstruct a single image~\cite{karpathypainting} or sample randomly in which case each network results in abstract patterns~\cite{ha2016abstract}. In contrast with these approaches, we propose to \emph{learn} the weights of this mapping network and condition it on a an image feature so that it can generate any image without image-specific weight optimization. Similarly, recent works have modeled 2D and 3D shapes using parametric and implicit functions~\cite{groueix2018papier,mescheder2019occupancy,park2019deepsdf,chen2019learning}. While previous attempts to apply this idea on images has focused on directly generating images on simple datasets such as MNIST~\cite{ha2016generating,chen2019learning}, we obtain a layer decomposition allowing various applications such as image editing and retrieval on complex images. 

\mypar{Deep, unsupervised, sequential image generation} 

We now present deep unupervised sequential approaches to image generation, the most related to our work.~\cite{Rolfe2013discriminative} uses a recurrent auto-encoder to reconstruct images iteratively, and employs a sparsity criterion to make sure that the image parts that are added at each iteration are simple. 
A second line of approaches~\cite{Gregor2015DRAW,Eslami2016AIR,gregor2016compression} are designed in a VAE framework.
Deep Recurrent Attentive Writer (DRAW)~\cite{Gregor2015DRAW} frames a recurrent approach using reinforcement learning and a spatial attention mechanism to mimic human gestures. A potential application of DRAW arises in its extension to conceptual image compression~\cite{gregor2016compression}, where a recurrent convolutional and hierarchical architecture allows to obtain various levels of lossy compressed images.
Attend, Infer, Repeat~\cite{Eslami2016AIR} models scenes by latent variables of object presence, content, and position. The parameters of presence and position are inferred by an RNN and a VAE decodes the objects one at a time to reconstruct images. 
A third strategy for learning sequential generative models is to  employ adversarial networks. Ganin et al.~\cite{ganin2018synthesizing} employ adversarial training in a reinforcement learning context. Specifically, their method dubbed SPIRAL, trains an agent to synthesize programs executed by a graphic engine to reconstruct images. 
The Layered Recursive GANs of~\cite{Yang2017LRGAN} learn to generate foreground and background images that are stitched together using STNs to form a consistent image. Although presented in a generic way that generalizes to multiple steps, the experiments are limited to foreground and background separation, made possible by the definition of a prior on the object size contained in the image. 
In contrast, our method (i) does not rely on STNs; (ii) extends to tens of steps as demonstrated in our experiments; (iii) relies on  simple architectures and losses, without the need of LSTMs or reinforcement learning.

\mypar{Image manipulation} 
Some successful applications of deep learning to image manipulation have been demonstrated, but they are usually specialized and offer limited user interaction. Image colorization \cite{zhang2016colorful} and style transfer \cite{Gatys2016ImageStyleTransfer} are two popular examples. 
Most approaches that allow user interaction are supervised. Zhu et al. \cite{Zhu2016manipulation} integrate user constraints in the form of brush strokes in GAN generations. More recently, Park et al. \cite{park2019SPADE} use semantic segmentation layouts and brush strokes to allow users to create new landscapes. 
In a similar vein, \cite{bau2018gan} locates sets of neurons related to different visual components of images, such as trees or artifacts, and allows their removal interactively. 
Approaches specialized in face editing, such as \cite{shen2016automatic} and \cite{portenier2018faceshop} demonstrate the large set of photo-realistic image manipulations that can be done to enhance quality, for instance background removal or swapping, diverse stylization effects, changes of the depth of field of the background, etc. 
These approach typically require precise label inputs from users, or training on heavily annotated datasets. Our approach provides an unsupervised alternative, with similar editing capacities.

\begin{figure*}[t!]
\begin{center}
\includegraphics[width=0.95\linewidth]{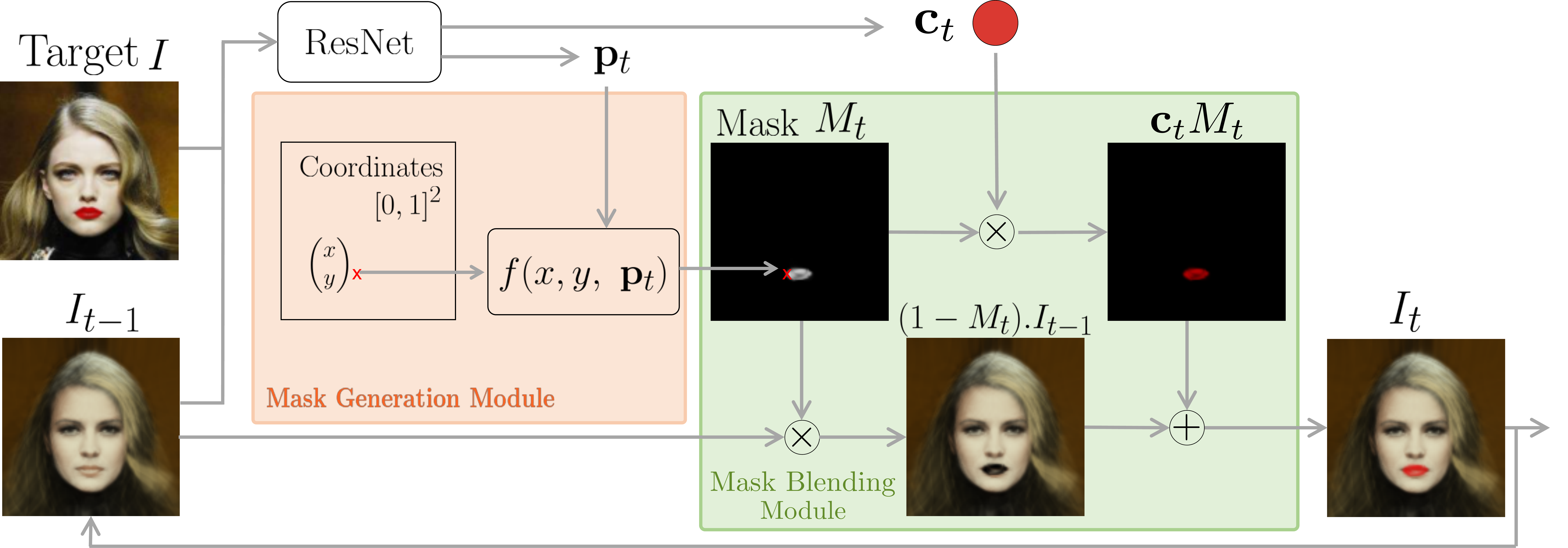}
\end{center}
    \vspace{-3mm}
    \caption{Our iterative generation pipeline for image reconstruction of target $I$. The previous canvas $I_{t-1}$ ($I_0$ can be initialized to a random uniform color) is concatenated with $I$ and forwarded through a ResNet feature extractor, to obtain a color $\textbf{c}_t$ and mask parameters $\textbf{p}_t$. 
    A Multi Layer Perceptron $f$ generates a parametric mask $M_t$ from pixelwise coordinates of a 2D grid and mask parameters $\textbf{p}_t$. 
    Our Mask Blending Module (in green) finally blends this mask with its corresponding color to the previous output $I_{t-1}$.}
\label{fig:mask_generation_and_blending}
\vspace{-2ex}
\end{figure*}

\section{Layered Vector Image Generation}

We frame image generation as an alpha-blending composition of a sequence of layers starting from a canvas of random uniform color $I_0$. Given a fixed budget of $T$ iterations, we iteratively blend $T$ generated colored masks onto the canvas. 
In this section, we first present our new architecture for vector image generation, then the training loss and finally discuss the advantages of our new architecture compared to existing approaches.

\subsection{Architecture}
The core idea of our approach is visualized in Fig. \ref{fig:mask_generation_and_blending}. At each iteration $t\in \left\lbrace 1...T\right\rbrace$, our model takes as input the concatenation of the target image $I \in \mathbb{R}^{3\times W\times H}$ and the current canvas $I_{t}$, and iteratively blends colored masks on the canvas resulting in $I_{t}$:

\begin{equation}
I_{t} = g(I_{t-1}, I),
\label{eq:aeset_k}
\end{equation} where $g$ consists of:
\begin{enumerate}[label=(\roman*)] 
\item a Residual Network (ResNet) that predicts mask parameters $\mathbf{p}_t \in \mathbb{R}^P$, with the corresponding color triplet $\mathbf{c}_t \in \mathbb{R}^3$, 
\item a mask generator module $f$, which generates an alpha-blending mask $M_t$ from the parameters $\mathbf{p}_t$, and 
\item our mask blending module that blends the masks $M_t$ with their color $\mathbf{c}_t$ on the previous canvas $I_{t-1}$.
\end{enumerate}

We represent the function $f$ generating the mask $M_t$ from $\mathbf{p}_t$ as a standard Multi-Layer Perceptron (MLP), which takes as input the concatenation of the mask parameters $\mathbf{p}_t$ and the two spatial coordinates $(x,y)$ of a point in image space. This MLP $f$ defines the continuous 2D function of the mask $M_t$ by:

\begin{equation}
    M_t(x,y) = f(x,y, \mathbf{p}_t).
\end{equation}

In practice, we evaluate the mask at discrete spatial locations corresponding to the desired resolution to produce a discrete image. We then update $I_t$ at each spatial location $(x,y)$ using the following blending:
\begin{equation}
I_{t}(x,y) = I_{t-1}(x,y).(1-M_{t}(x,y)) + \mathbf{c}_t.M_{t}(x,y),
\label{alpha_blending}
\end{equation}

where $I_{t}(x,y)\in \mathbb{R}^3$ is the RGB value of the resulting image $I_t$ at position $(x,y)$. We note that, at test time, we may perform a different number of iterations $N$ than the one during training $T$. Choosing $N>T$ may help to model accurately images that contain complex patterns, as we show in our experiments.  

All the design choices of our approach are justified in detail in Section \ref{ssec:discussion} and supported empirically by experiments and ablations in Section \ref{sec:choices}.

\subsection{Training losses}

We learn the weights of our network end-to-end by minimizing a reconstruction loss between the target $I$ and our result $R= I_T$. We perform experiments either using an $\ell_1$ loss, which enables simple quantitative comparisons, or a perceptual loss \cite{Johnson2016Perceptual}, leading to visually improved results. Our perceptual loss $\mathcal{L}_{perc}$ is based on the Euclidean norm $\left\lVert . \right\rVert_2$ between feature maps $\phi(.)$ extracted from a pre-trained VGG16 network and the Frobenius norm between the Gram matrices obtained from these feature maps $G(\phi(.))$:

$$\mathcal{L}_{perc} = \mathcal{L}_{content} + \lambda \mathcal{L}_{style},$$
where $$\mathcal{L}_{content}(I,R) = \left\lVert \phi(I)- \phi(R) \right\rVert_2,$$
$$\mathcal{L}_{style}(I,R) = \left\lVert G(\phi(I)) - G(\phi(R)) \right\rVert_F,$$
and $\lambda$ is a non-negative scalar that controls the relative influence of the style loss. 
To obtain even sharper results, we may optionally add an adversarial loss. In this case, a discriminator $D$ is trained to recognize real images from generated ones, and we optimize our generator $G$ to fool this discriminator. We train $D$ to minimize the non saturating GAN loss from~\cite{goodfellow2014generative} with R1 Gradient Penalty loss~\cite{mescheder2018training}.
The architecture of $D$ is the patch discriminator defined in~\cite{Isola2016ImageToImage}.

\subsection{Discussion} \label{ssec:discussion}
\paragraph{Architecture choices.}

Our architecture choices are related to desirable properties of the final generation model:

\textit{\textbf{Layered decomposition:}} This choice allows us to obtain a mask decomposition which is a key component of image editing pipelines.
Defining one color per layer, similar to image simplification and quantization approaches, is important to obtain visually coherent regions. We further show that a single layer baseline does not perform as well.

\textit{\textbf{Vectorized layers:}} By using a lattice input for the mask generator, it is possible to perform local image editing and generation at any resolution without introducing up-sampling artifacts or changing our model architecture. This vector mask representation is especially convenient for HD image editing.  

\textit{\textbf{Recursive vs one-shot:}} We generate the mask parameters recursively to allow the model to better take into account the interaction between the different masks. We show that a one-shot baseline, where all the mask parameters are predicted in a single pass leads to worse results. Moreover, as mentioned above and demonstrated in the experiments, our recursive procedure can be applied a larger number of times to model more complex images.

\vspace{-2ex}
\paragraph{Number of layers vs. size of the mask parameter.} Our mask blending module iteratively adds colored masks to the canvas to compose the final image. The size of the mask parameter $\mathbf{p}$ controls the complexity of the possible mask shapes, while the number of masks controls the amount of different shapes that be used to compose the image. Since we aim at producing a set of layers that can easily be used and interpreted by a human, we use a limited number of strokes and masks. 

\vspace{-2ex}
\paragraph{Complexity of the mask generator network.} Interestingly, if the network generating the masks from the parameters was very large, it could generate very complex patterns. In fact, one could show using the universal approximation theorem \cite{cybenko1989approximation,hornik1991approximation} that, with a large number of hidden units in the MLP $f$, an image could be approximated with only three layers ($N=3$) of our generation process, using one mask for each color channel. Thus it is important to control the complexity of $f$ to obtain meaningful primitive shapes. For example, we found that replacing our MLP by a ResNet leads to less interpretable masks (see Section \ref{sec:choices} and Fig.~\ref{fig:iter_resnet_masks}).

\section{Experiments}

\begin{figure*}[htb]
    \centering
    \includegraphics[width=\linewidth]{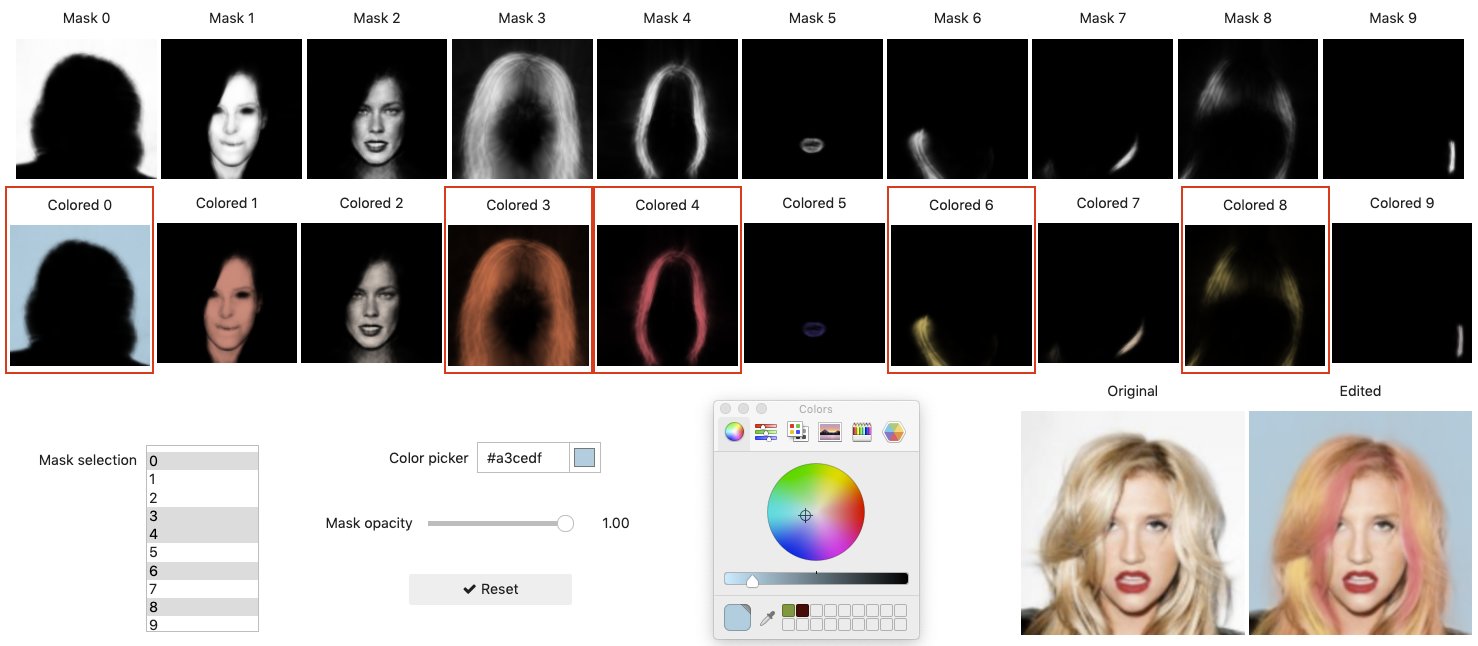}
    \caption{Our editing interface using automatically extracted masks to bootstrap the editing process.}
    \label{fig:editing_interface}
\end{figure*}

In this section we first introduce the datasets, the training and network architecture details, then we demonstrate the practical interest of our approach in several applications, and justify the architecture choices in extensive ablation studies.

\subsection{Datasets and implementation details}

\paragraph{Datasets.}
Our models are trained on two datasets, CelebA~\cite{celebA} (202k images of celebrity faces) and ImageNet~\cite{imagenet_cvpr09} (1.28M natural images of 1000 classes), using images downsampled to $128\times128$.

\paragraph{Training details.}

The parameters of our generator $g$ are optimized using Adam~\cite{kingma2014adam} with a learning rate of $2\times10^{-4}$, $\beta_1=0.9$ and no weight decay. The batch size is set to 32 and training image size is fixed to $128\times128$ pixel images.

\begin{figure*}[htb]
  \centering
  \begin{subfigure}{0.33\linewidth}
        \includegraphics[width=0.3\linewidth]{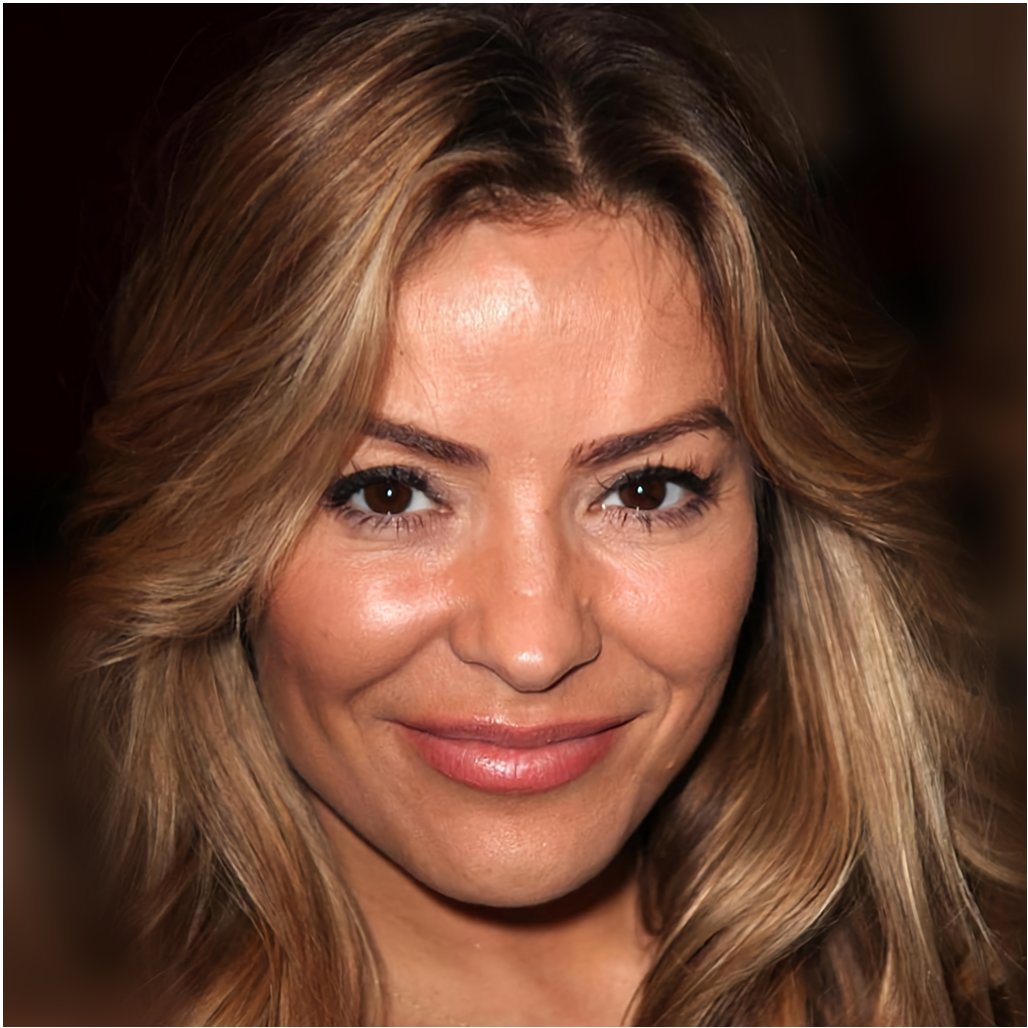}
        \includegraphics[width=0.3\linewidth]{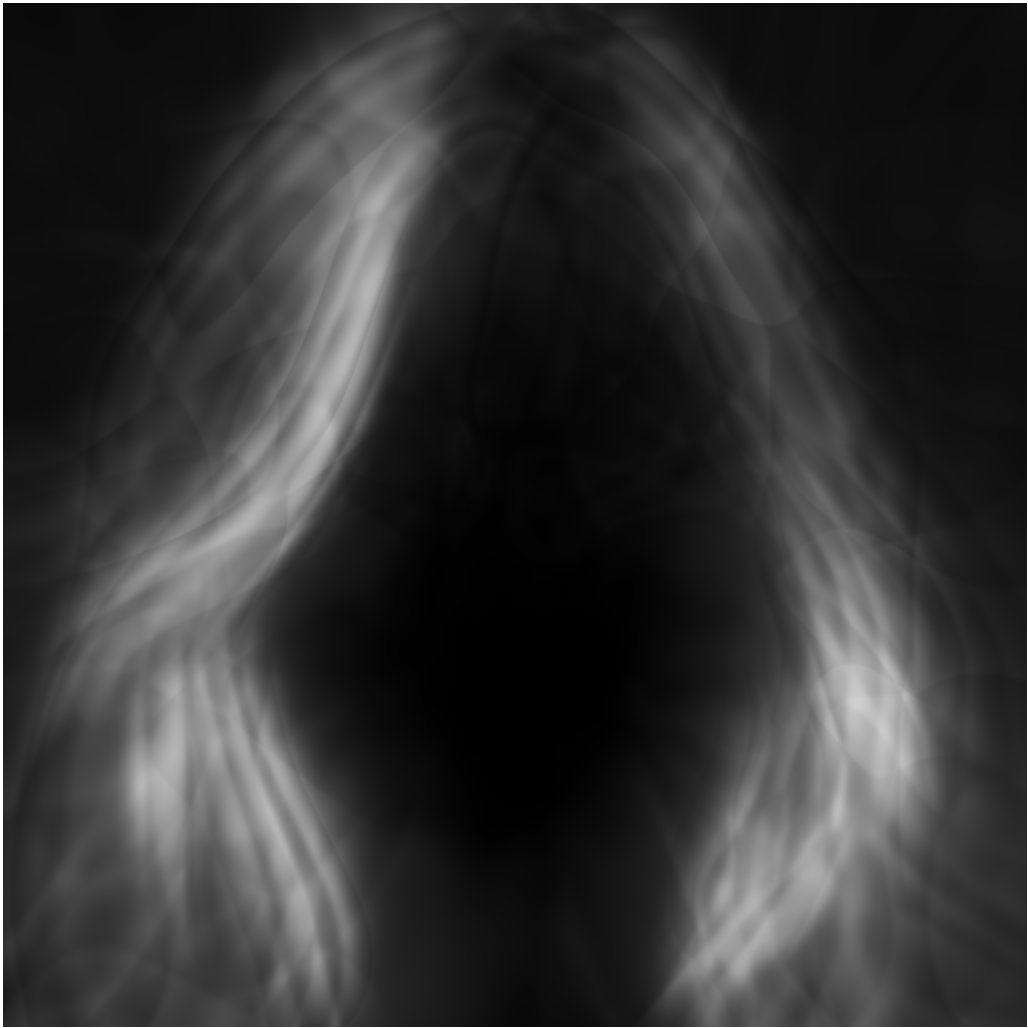}
        \includegraphics[width=0.3\linewidth]{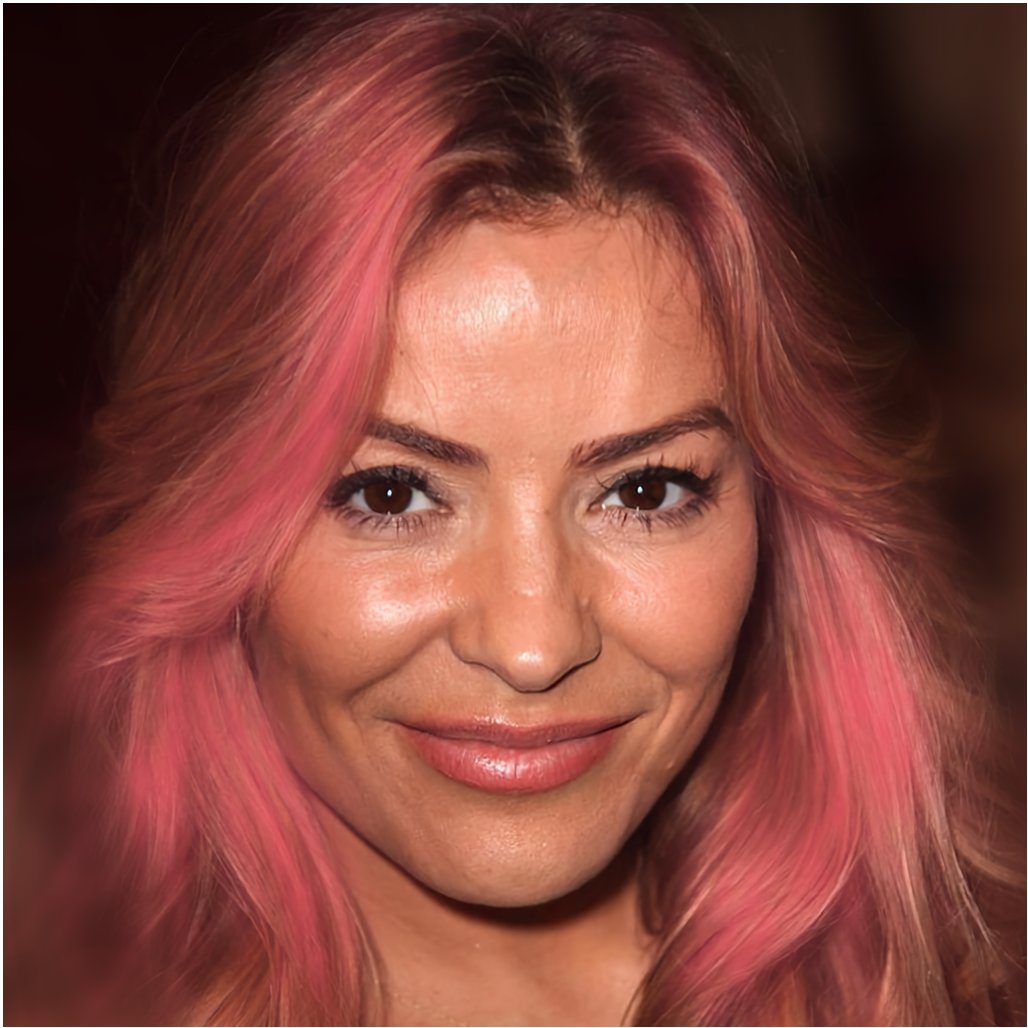}\\
        \includegraphics[width=0.3\linewidth]{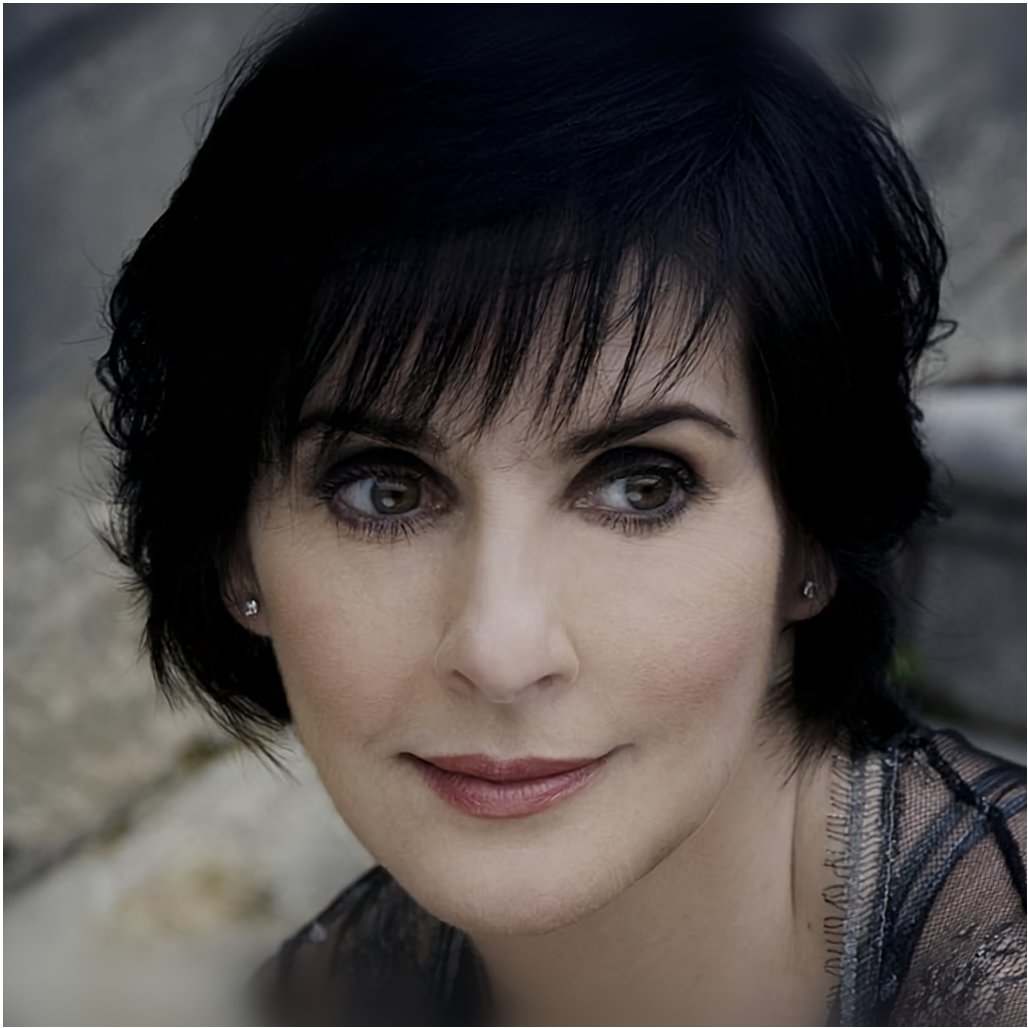}
        \includegraphics[width=0.3\linewidth]{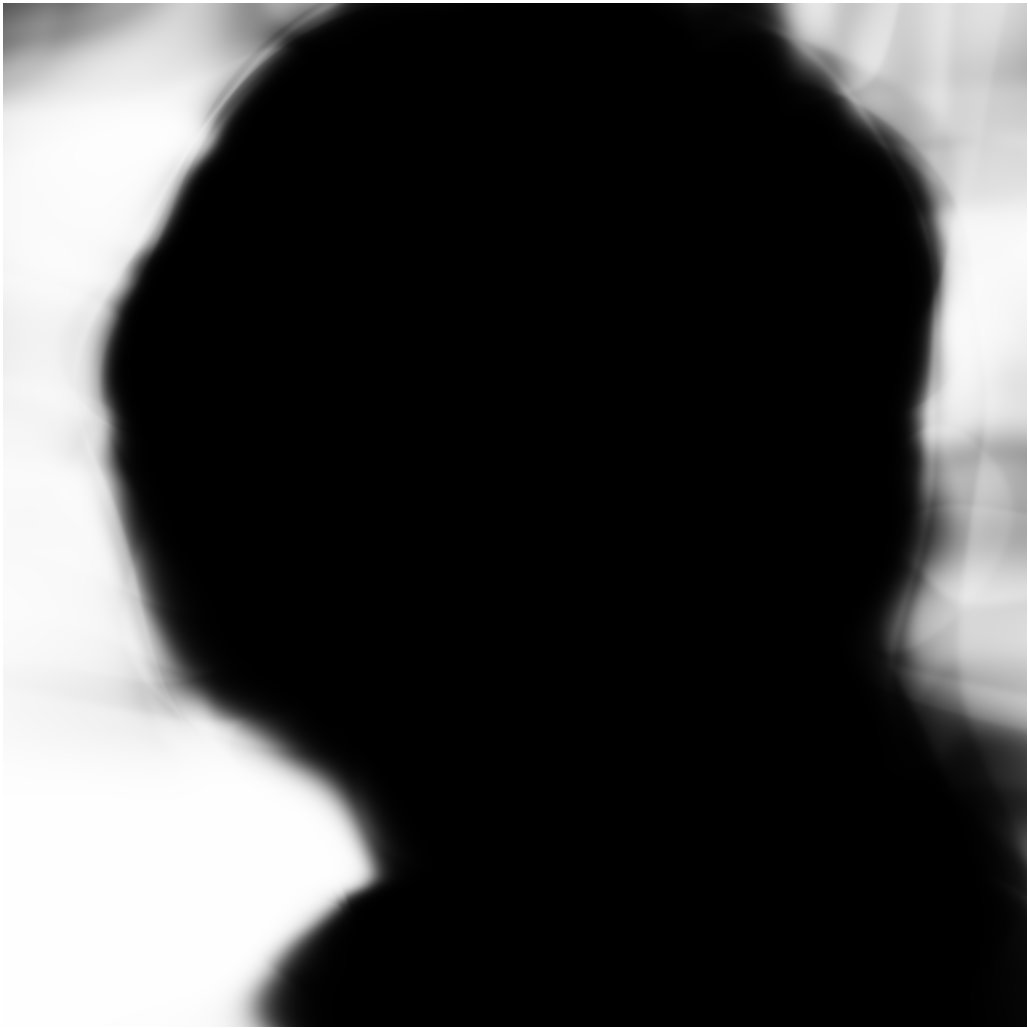}
        \includegraphics[width=0.3\linewidth]{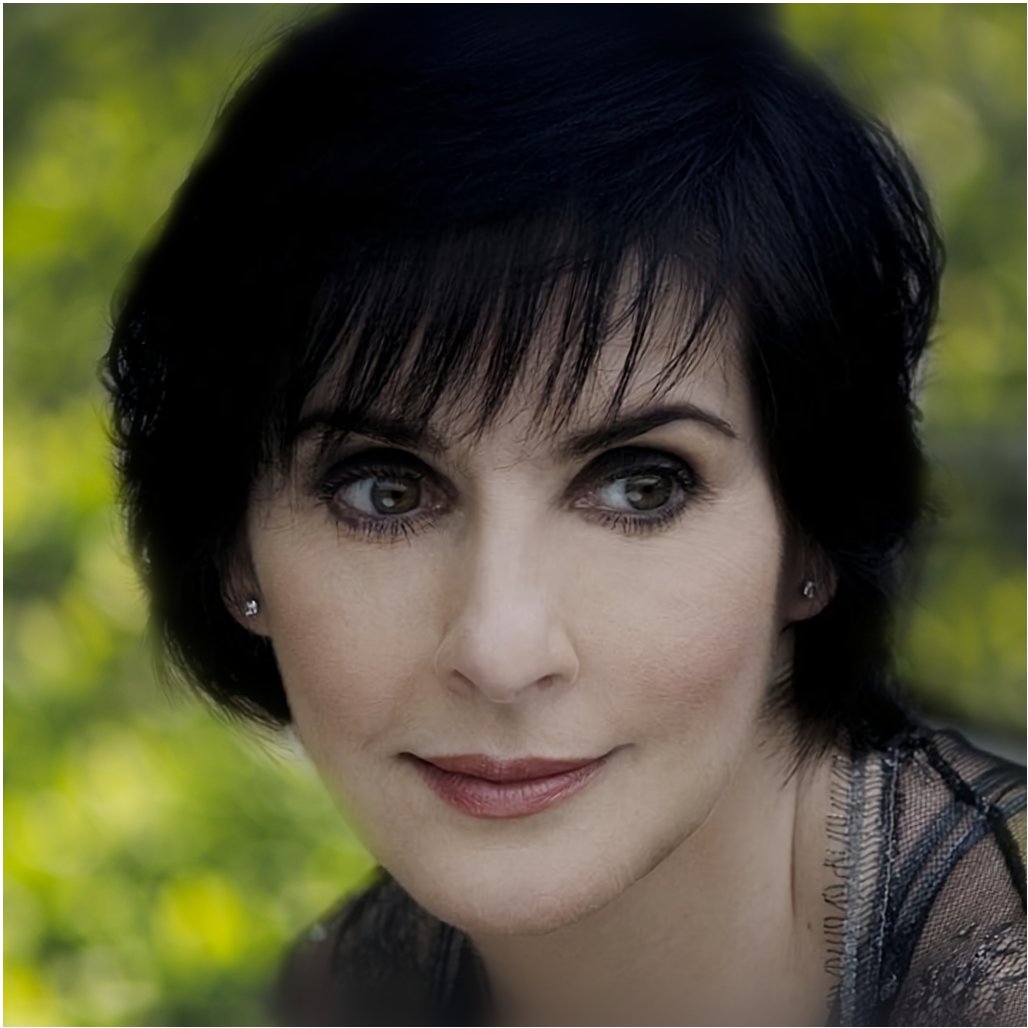}\\
        \includegraphics[width=0.3\linewidth]{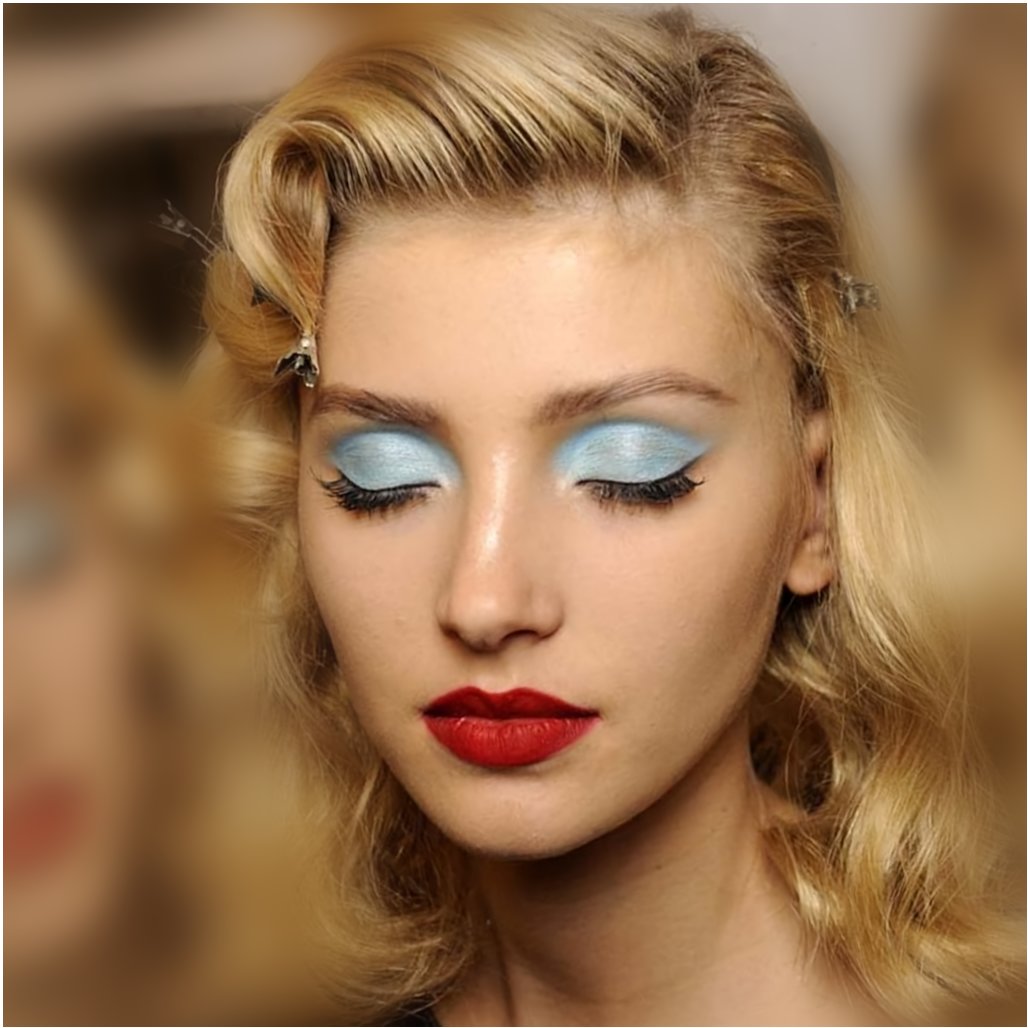}
        \includegraphics[width=0.3\linewidth]{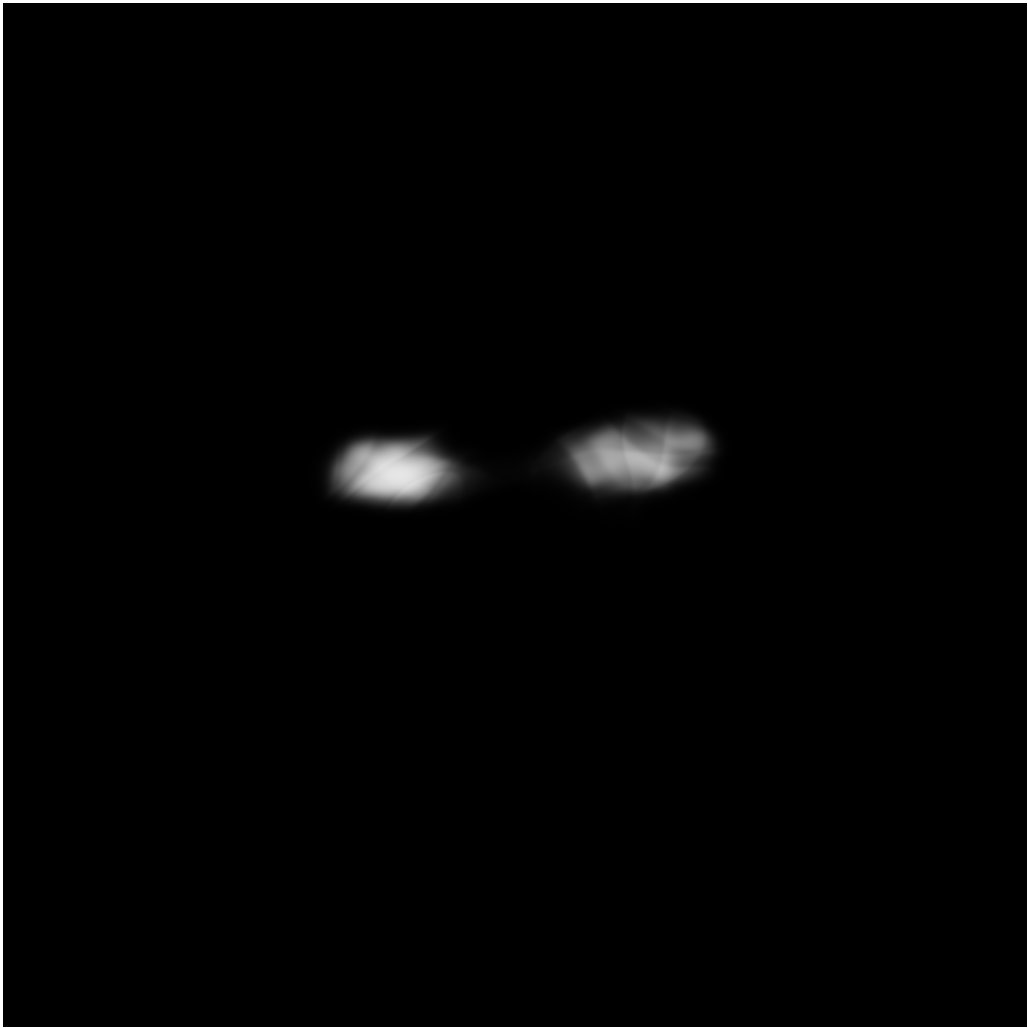}
        \includegraphics[width=0.3\linewidth]{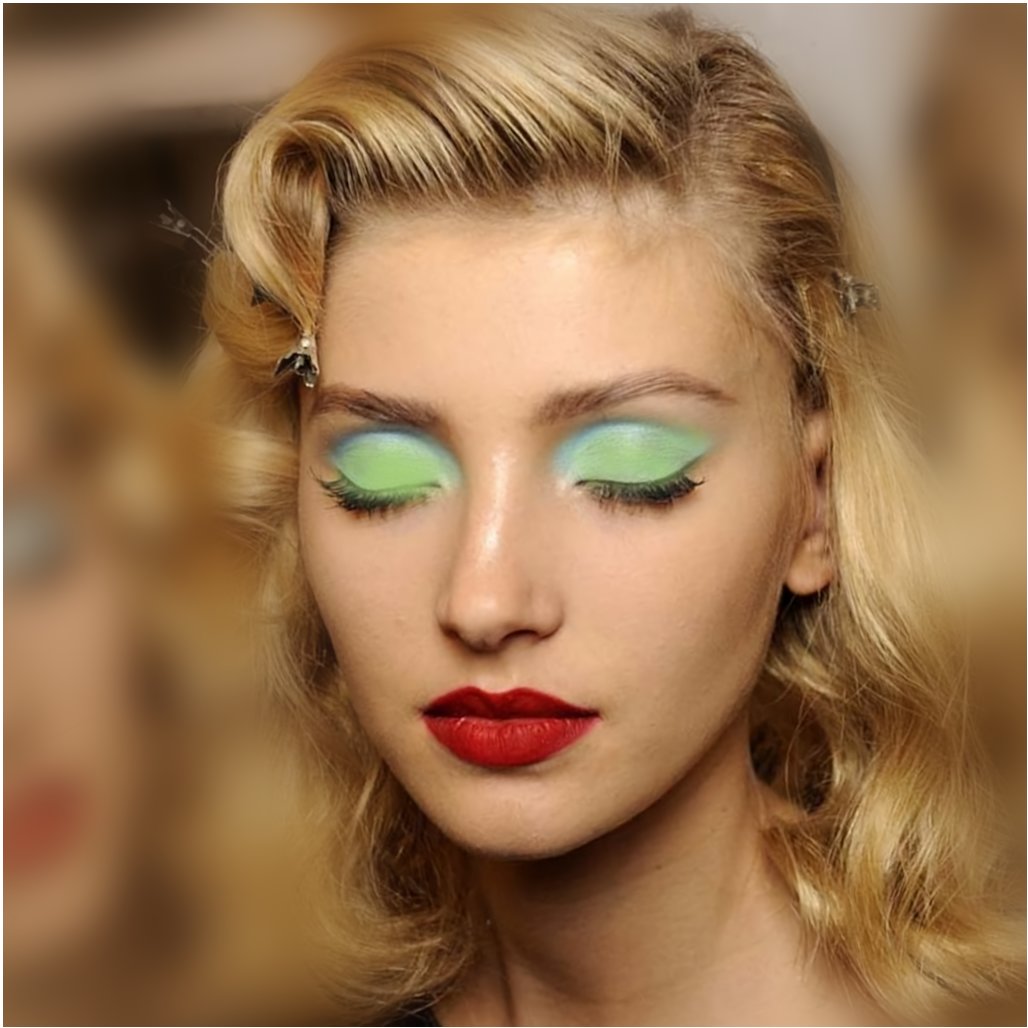} 
        \end{subfigure}~    
  \begin{subfigure}{0.33\linewidth}
       \includegraphics[width=0.3\linewidth]{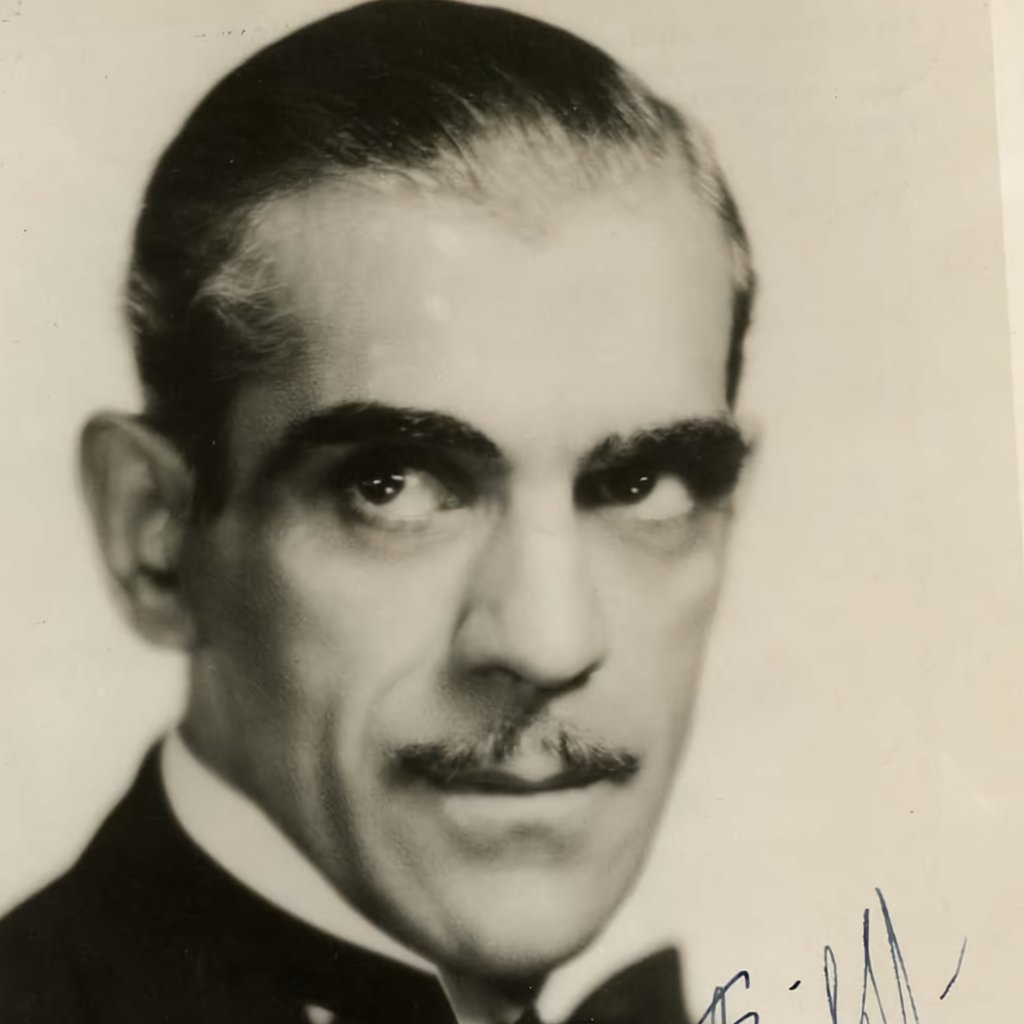}
         \includegraphics[width=0.3\linewidth]{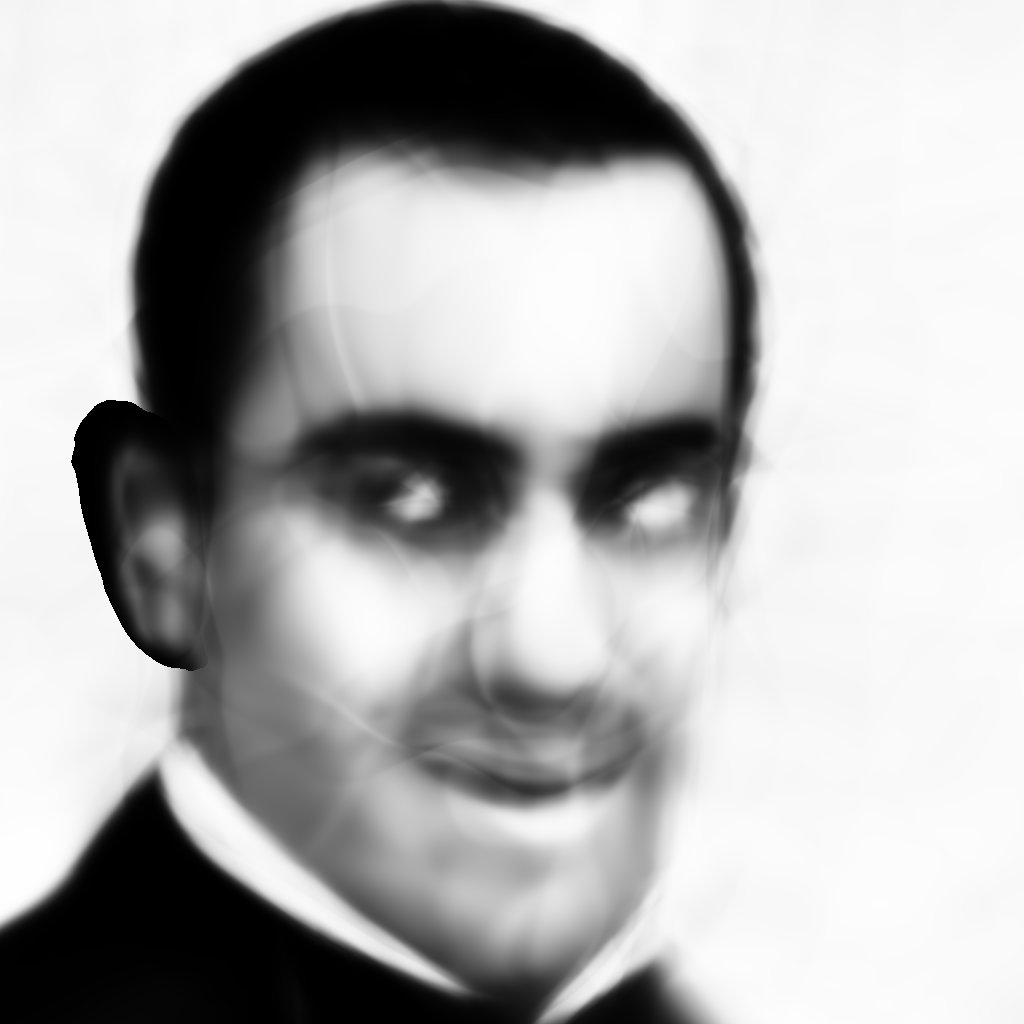}
         \includegraphics[width=0.3\linewidth]{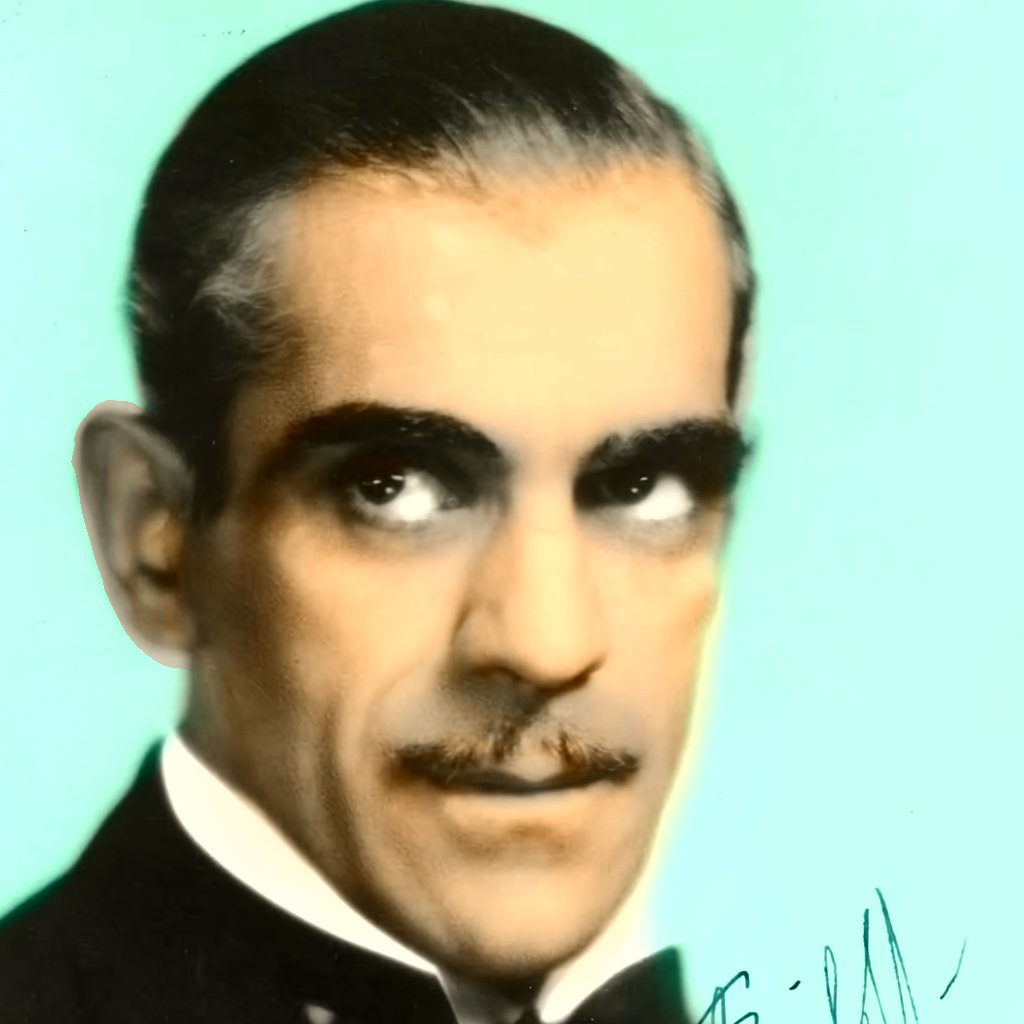}\\
        \includegraphics[width=0.3\linewidth]{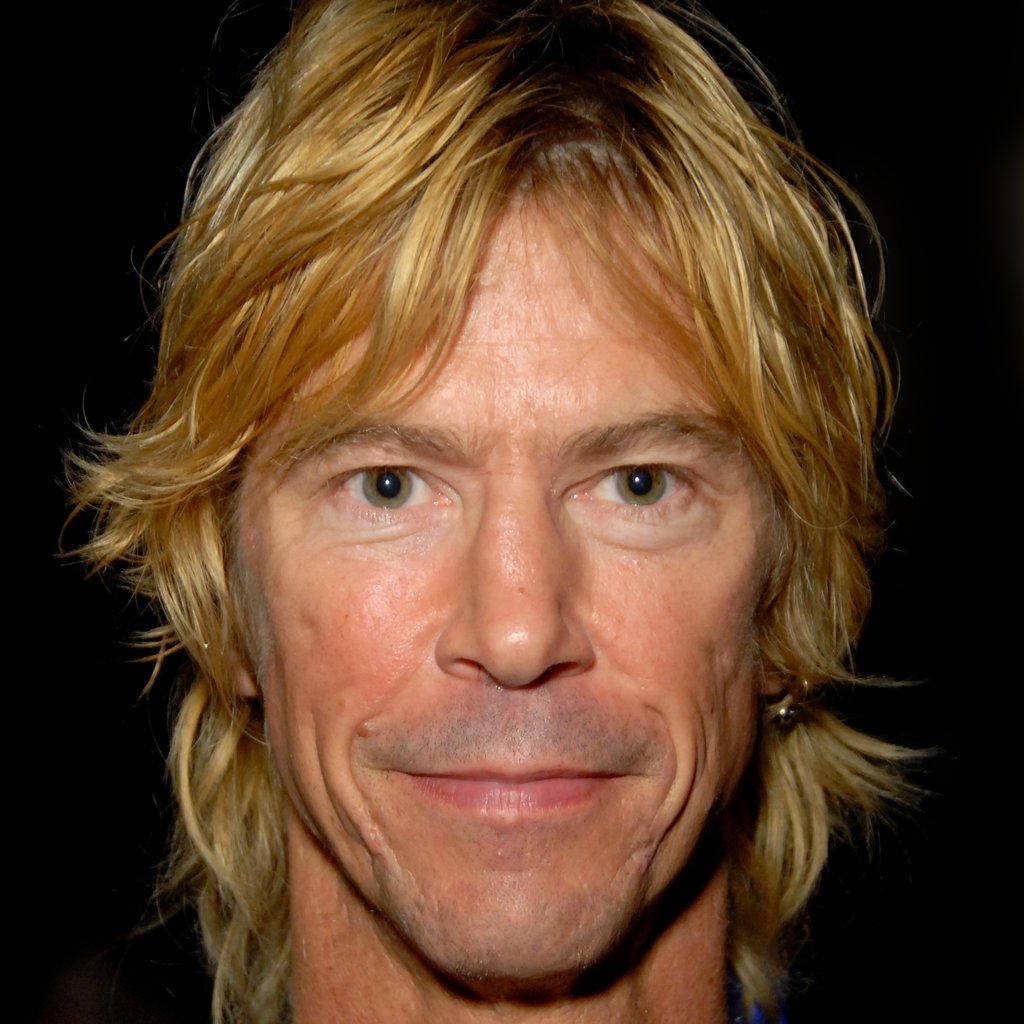}
        \includegraphics[width=0.3\linewidth]{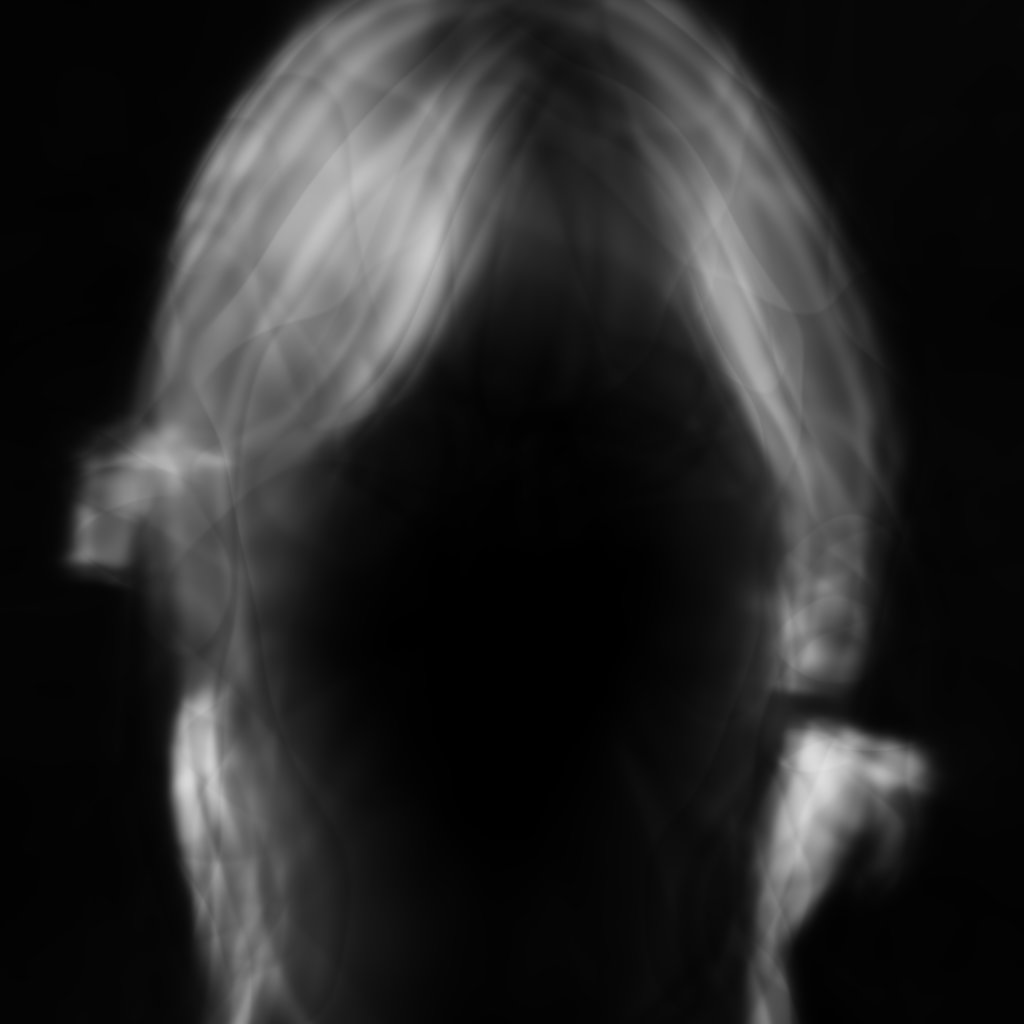}
        \includegraphics[width=0.3\linewidth]{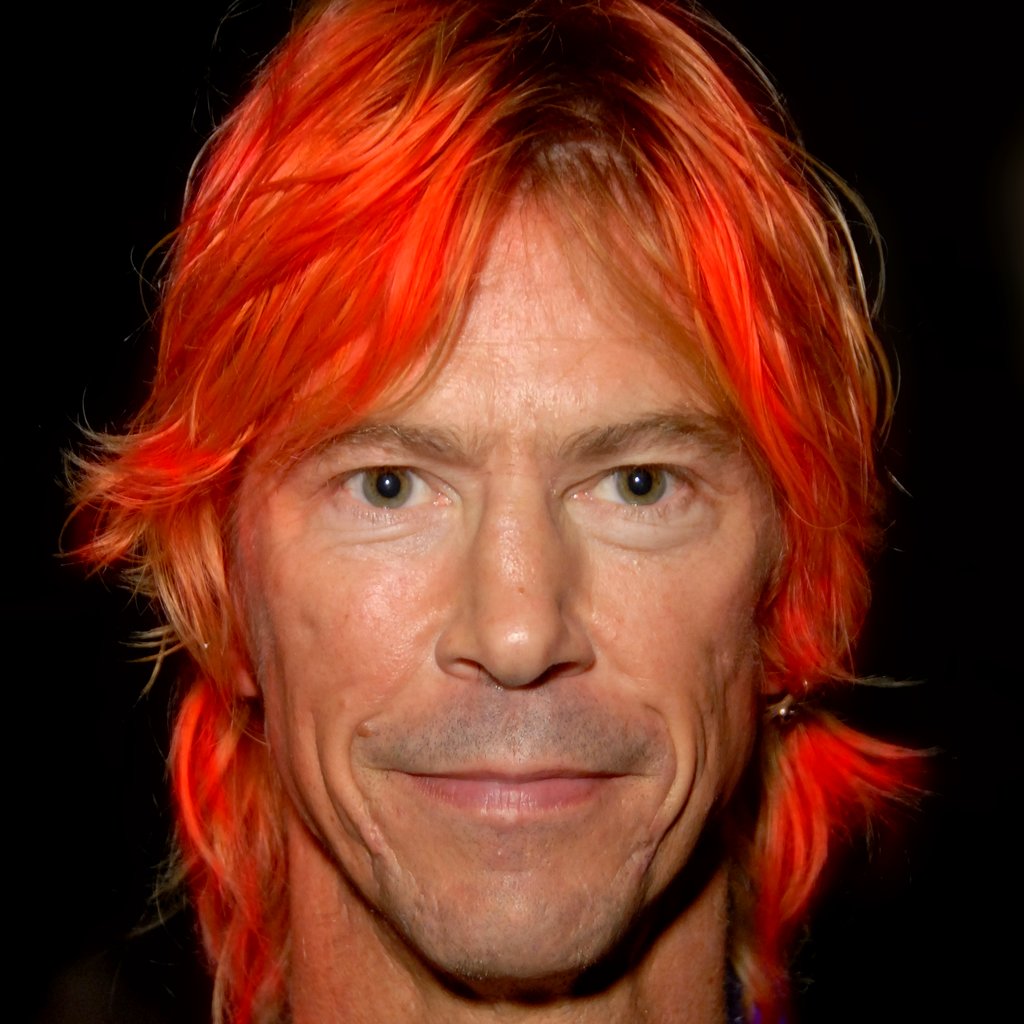}\\
        \includegraphics[width=0.3\linewidth]{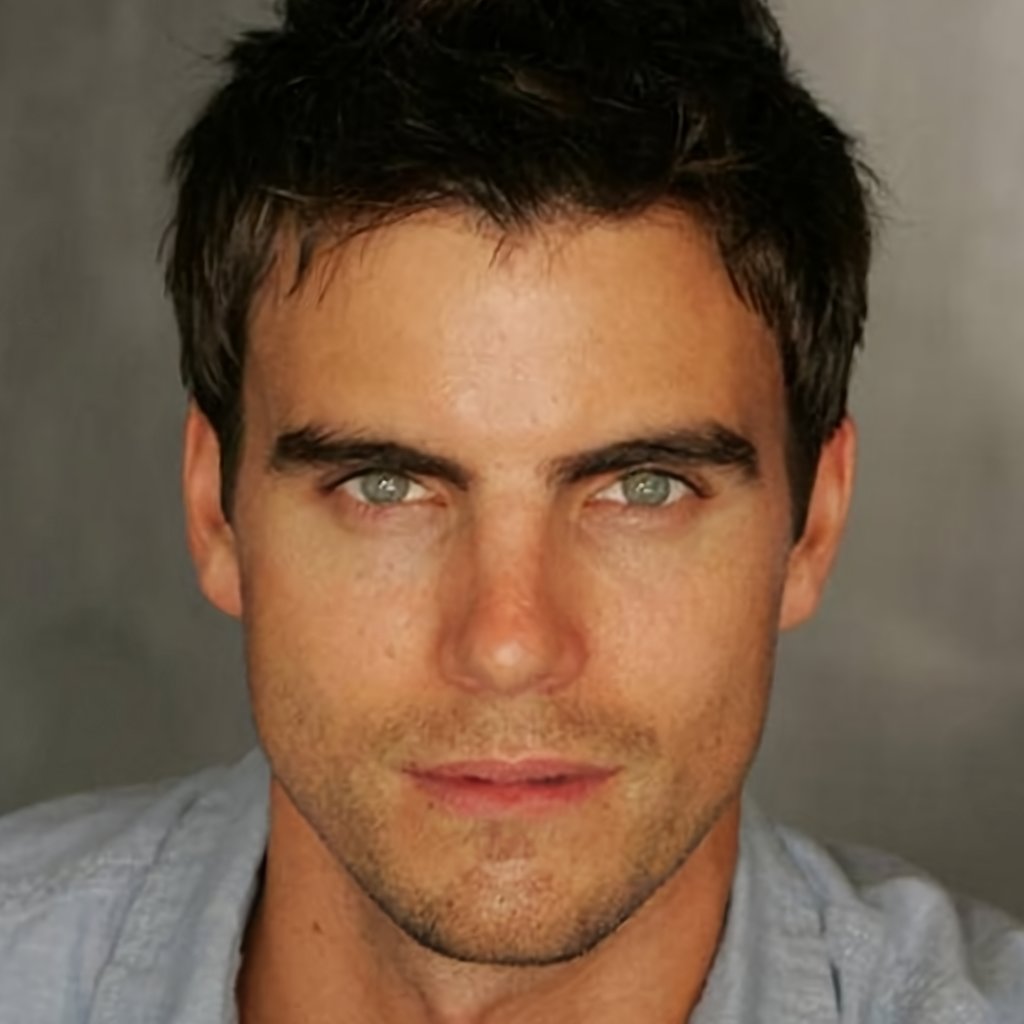}
        \includegraphics[width=0.3\linewidth]{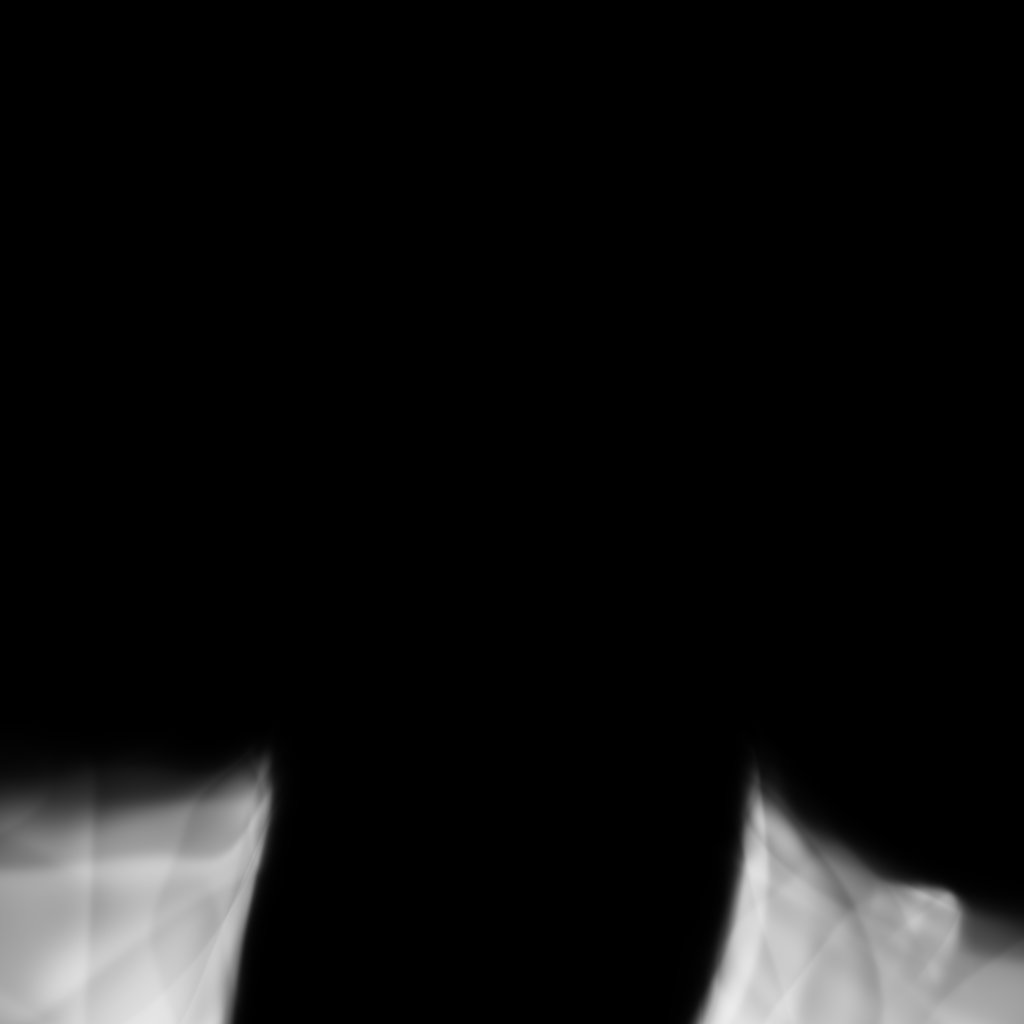}
        \includegraphics[width=0.3\linewidth]{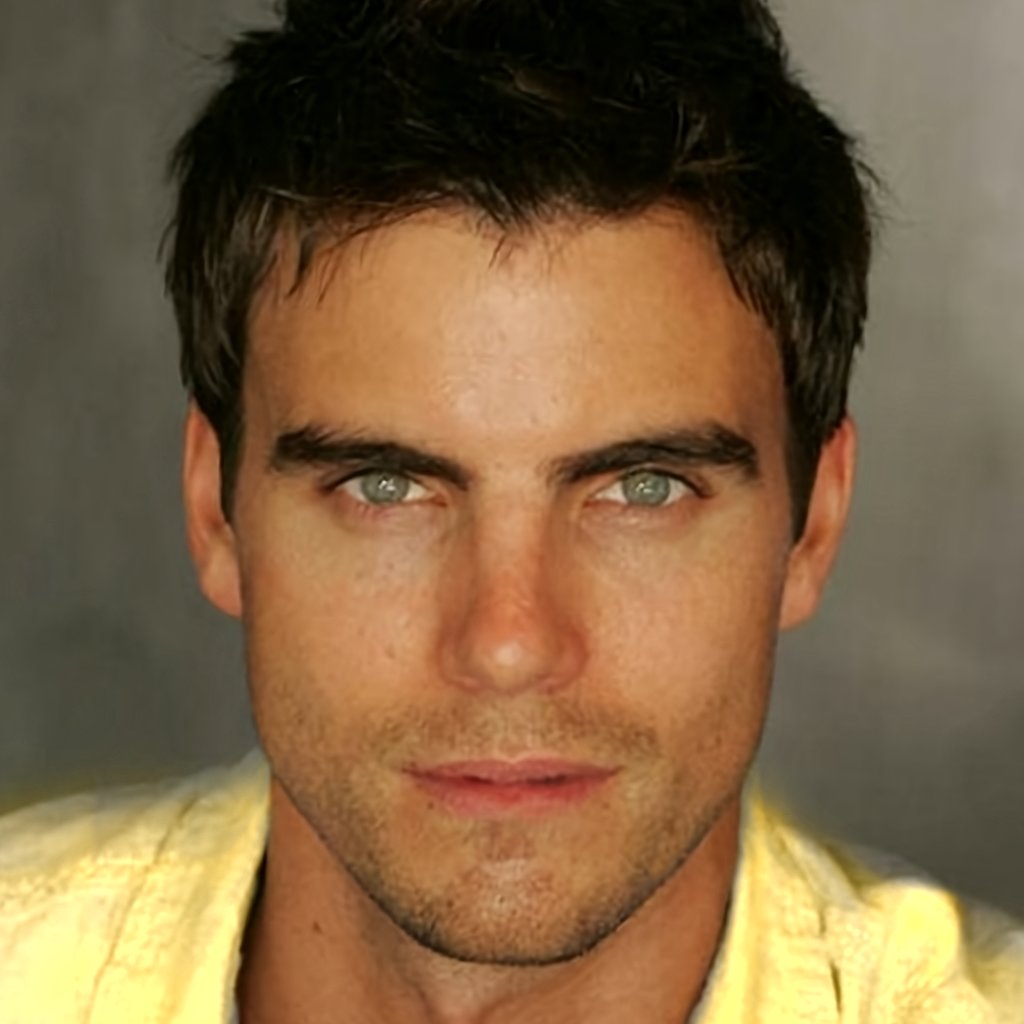}
    \end{subfigure}~~~
  \begin{subfigure}{0.33\linewidth}
            \includegraphics[width=0.3\linewidth]{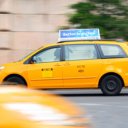}
            \includegraphics[width=0.3\linewidth]{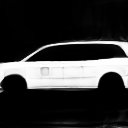}
            \includegraphics[width=0.3\linewidth]{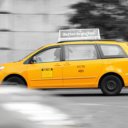}\\
            \begin{minipage}[c]{0.3\linewidth}
                \includegraphics[width=1\linewidth]{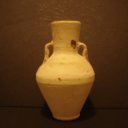}
            \end{minipage}
            \begin{minipage}[c]{0.3\linewidth}
                \includegraphics[width=0.45\linewidth]{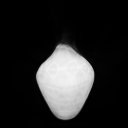}
                \includegraphics[width=0.45\linewidth]{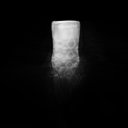}\\
                \includegraphics[width=0.45\linewidth]{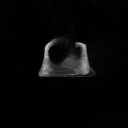}
                \includegraphics[width=0.45\linewidth]{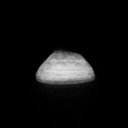}
            \end{minipage}
            \begin{minipage}[c]{0.3\linewidth}
            \includegraphics[width=1\linewidth]{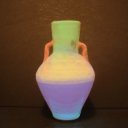}
            \end{minipage}\\
            \includegraphics[width=0.3\linewidth]{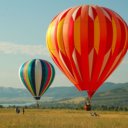}
            \includegraphics[width=0.3\linewidth]{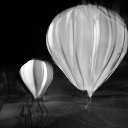}
            \includegraphics[width=0.3\linewidth]{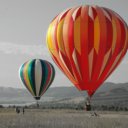}
     \end{subfigure}
    \caption{Some editings on CelebA and ImageNet, using little supervision (mask selection in one click and new style/color selection). Note that the CelebA editings are performed on $1024\times 1024$ images. Left: original; center: mask; right: edit.}
    \label{fig:editing_examples_Imagenet_CelebA}
    \vspace{-4mm}
\end{figure*}

\begin{figure}[htb]
    \centering
    \includegraphics[width=\linewidth]{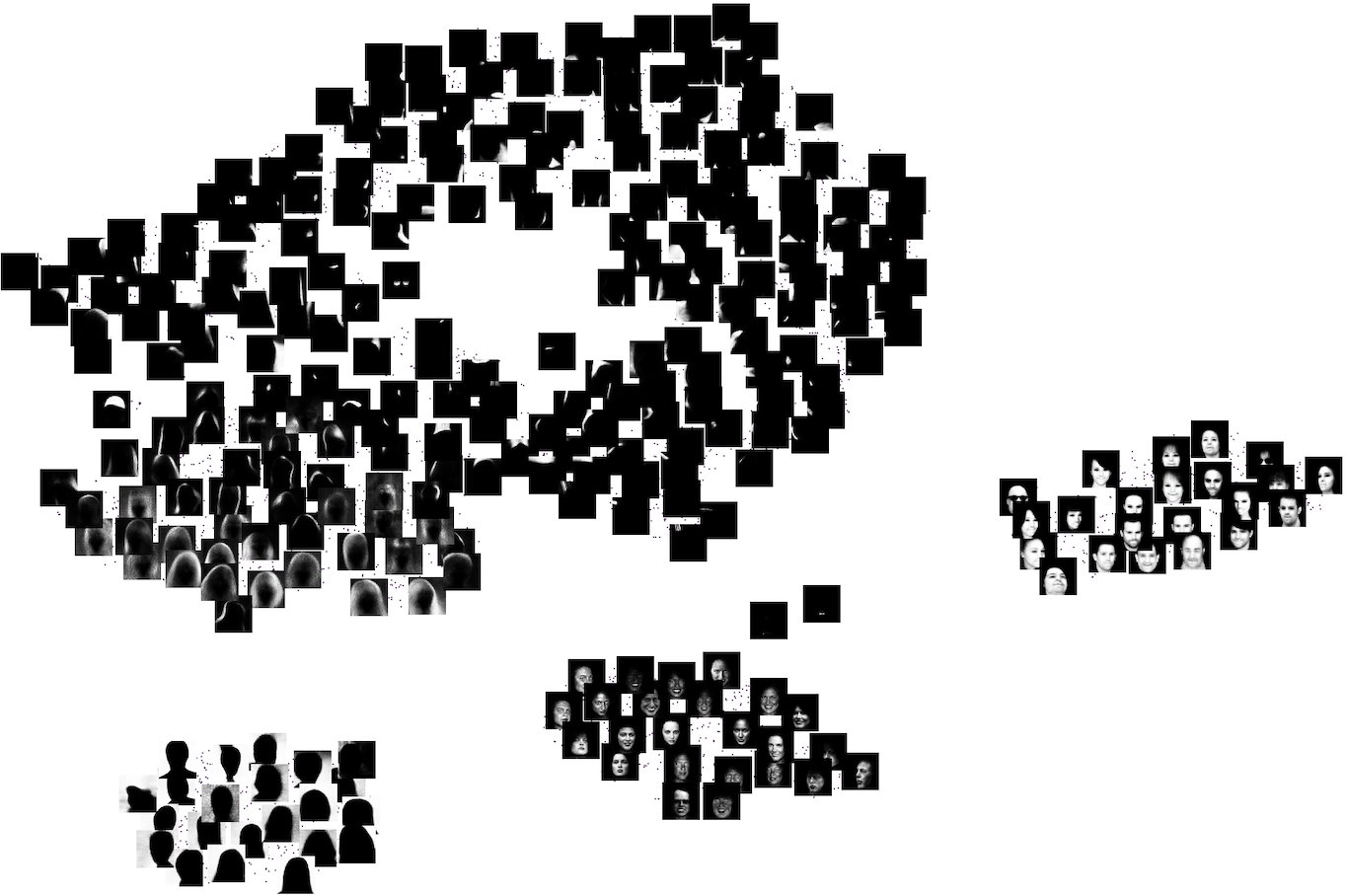}
    \caption{t-SNE visualization of masks obtained from 5000 reconstructions on CelebA.}
    \label{fig:tsne}
\end{figure}

\paragraph{Network architectures.}
The mask generator $f$ consists of an MLP with three hidden layers of 128 units with group normalization~\cite{wu2018group}, tanh non-linearities, and an additional sigmoid after the last layer. $f$ takes as input a parameter vector $\mathbf{p}$ and pixel coordinates $(x,y)$, and outputs a value between 0 and 1. The parameter $\mathbf{p}$ and the color $\mathbf{c}$ are predicted by a ResNet-18 network. Further details about the network architectures are in the supplementary material.

\subsection{Applications}

We now demonstrate how our image decomposition may serve different purposes such as image editing, retrieval and vectorization.

\paragraph{Image editing.}

Image editing from raw pixels can be time consuming. Using our generated masks, it is possible to alter the original image by applying edits such as luminosity or color modifications on the region specified by a mask. 
Fig.~\ref{fig:editing_interface} shows an interface we designed for such editing showing the masks corresponding to the image. It avoids going through the tedious process of defining a blending mask manually.  The learned masks capture the main components of the image, such as the background, face, hairs, lips. Fig.~\ref{fig:editing_examples_Imagenet_CelebA} demonstrate a variety of editing we performed and the associated masks.
Our approach works well on the CelebA dataset, and allows to make simple image modifications on the more challenging ImageNet images. To optimize our results on ImageNet, the edits of Fig.~\ref{fig:editing_examples_Imagenet_CelebA} are obtained by finetuning our model on images of each object class. 

\paragraph{Attribute-based image retrieval.}

A t-SNE \cite{maaten2008visualizing} visualization of the mask parameters obtained on CelebA is shown in Fig.~\ref{fig:tsne}. Different clusters of masks are clearly visible, for backgrounds, hairs, face shadows, etc. This experiment highlights the fact our approach naturally extract semantic components of face images.

\begin{figure}[htb]
    \centering
    \setlength{\tabcolsep}{1pt}
    \begin{tabular}{cccccc}
    
    \small{Hat} & \small{Shirt}   & \small{Backgrd} & \small{Glasses} & \small{Face}      & \small{Lipstick} \\
     \small{}  &  \small{collar} & \small{text}     &                 & \small{direction} &                  \\
    
    \includegraphics[width=0.15\linewidth]{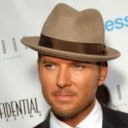}&
    \includegraphics[width=0.15\linewidth]{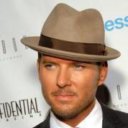}& 
    \includegraphics[width=0.15\linewidth]{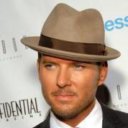}&
    \includegraphics[width=0.15\linewidth]{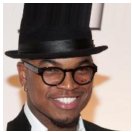}& 
    \includegraphics[width=0.15\linewidth]{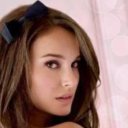}&
    \includegraphics[width=0.15\linewidth]{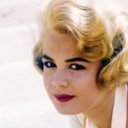}\\
    
    \includegraphics[width=0.15\linewidth]{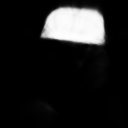}&
    \includegraphics[width=0.15\linewidth]{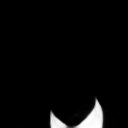}&
    \includegraphics[width=0.15\linewidth]{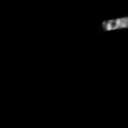}&
    \includegraphics[width=0.15\linewidth]{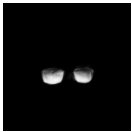}&
    \includegraphics[width=0.15\linewidth]{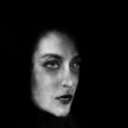}&
    \includegraphics[width=0.15\linewidth]{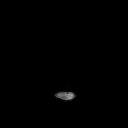}\\
    
    \includegraphics[width=0.15\linewidth]{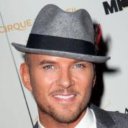}&
    \includegraphics[width=0.15\linewidth]{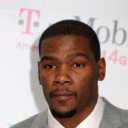}&
    \includegraphics[width=0.15\linewidth]{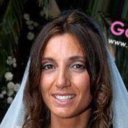}&
    \includegraphics[width=0.15\linewidth]{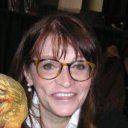}&
    \includegraphics[width=0.15\linewidth]{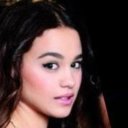}&
    \includegraphics[width=0.15\linewidth]{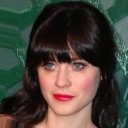}\\
    
    \includegraphics[width=0.15\linewidth]{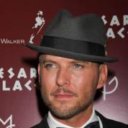}&
    \includegraphics[width=0.15\linewidth]{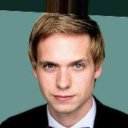}&
    \includegraphics[width=0.15\linewidth]{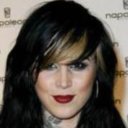}&
    \includegraphics[width=0.15\linewidth]{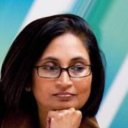}&
    \includegraphics[width=0.15\linewidth]{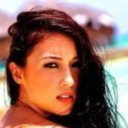}&
    \includegraphics[width=0.15\linewidth]{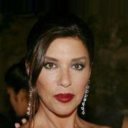}\\
    
    \includegraphics[width=0.15\linewidth]{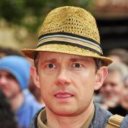}&
    \includegraphics[width=0.15\linewidth]{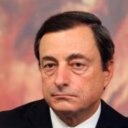}&
    \includegraphics[width=0.15\linewidth]{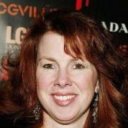}&
    \includegraphics[width=0.15\linewidth]{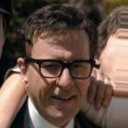}&
    \includegraphics[width=0.15\linewidth]{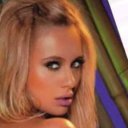}&
    \includegraphics[width=0.15\linewidth]{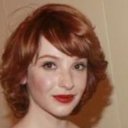}\\
    
    \includegraphics[width=0.15\linewidth]{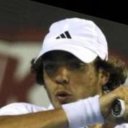}&
    \includegraphics[width=0.15\linewidth]{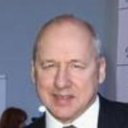}&
    \includegraphics[width=0.15\linewidth]{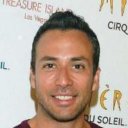}&
    \includegraphics[width=0.15\linewidth]{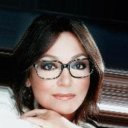}&
    \includegraphics[width=0.15\linewidth]{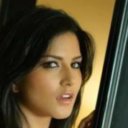}&
    \includegraphics[width=0.15\linewidth]{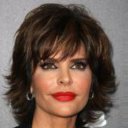}\\
    
    \includegraphics[width=0.15\linewidth]{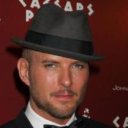}&
    \includegraphics[width=0.15\linewidth]{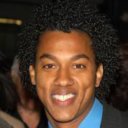}&
    \includegraphics[width=0.15\linewidth]{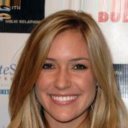}&
    \includegraphics[width=0.15\linewidth]{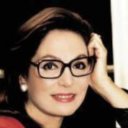}&
    \includegraphics[width=0.15\linewidth]{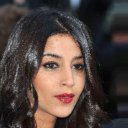}&
    \includegraphics[width=0.15\linewidth]{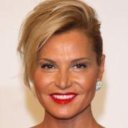}\\
    
    \includegraphics[width=0.15\linewidth]{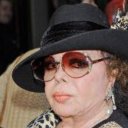}&
    \includegraphics[width=0.15\linewidth]{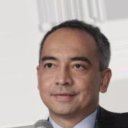}&
    \includegraphics[width=0.15\linewidth]{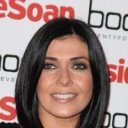}&
    \includegraphics[width=0.15\linewidth]{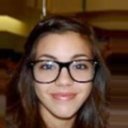}&
    \includegraphics[width=0.15\linewidth]{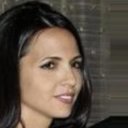}&
    \includegraphics[width=0.15\linewidth]{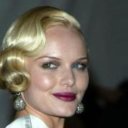}\\
    \end{tabular}
    \vspace{-2mm}
    \caption{Given a target image and a mask of an area of interest extracted from it, a nearest neighbor search in the learned mask parameter space allows the retrieval of images sharing the desired attribute with the target.}
    \label{fig:nn_mask_par}
\end{figure}

Our approach may be used in an image search content: given a query image, a user can select a mask that displays a particular attribute of interest and search for images which decomposition includes similar masks. Suppose we would like to retrieve pictures of people wearing a hat as  displayed in a query image, we can easily extract the mask that corresponds to the hat in our decomposition and its parameters. Nearest neighbor for different masks, using a cosine similarity distance between mask parameters $\mathbf{p}$ are provided in Fig.~\ref{fig:nn_mask_par}. Note how different masks extracted from the same query image lead to very different retrieved images. Such a strategy could potentially be used for efficient image annotation or few-shot learning. We evaluated oneshot nearest neighbor classification for the "Wearing Hat" and "Eyeglasses" categories in CelebA using the hat and glasses examples shown in Fig.~\ref{fig:nn_mask_par}, and obtained respectively $34\%$ and $49\%$ average precision. Results for eyeglasses attribute were especially impressive with $33\%$ recall at $98\%$ precision, compared to a low recall (less than $10\%$ at $98\%$ precision) for a baseline using cosine distance between features of a Resnet18 trained on ImageNet.

\vspace{-2.5mm}
\paragraph{Vector image generation.}

\begin{figure}[htb]
    \centering
    \begin{tabular}{ccc}
     Original  &Bilinear&  Ours\\ 
     \includegraphics[width=0.25\linewidth]{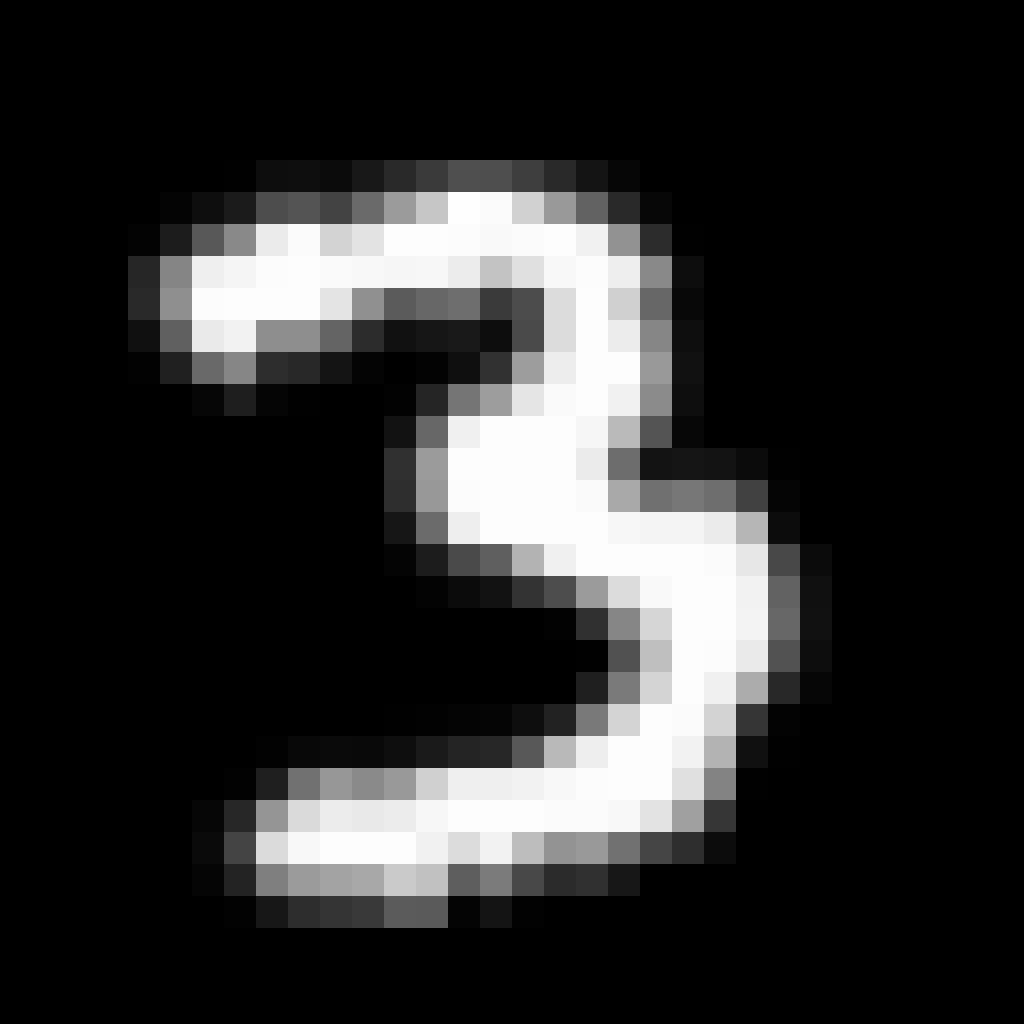}&
    
            \includegraphics[width=0.25\linewidth]{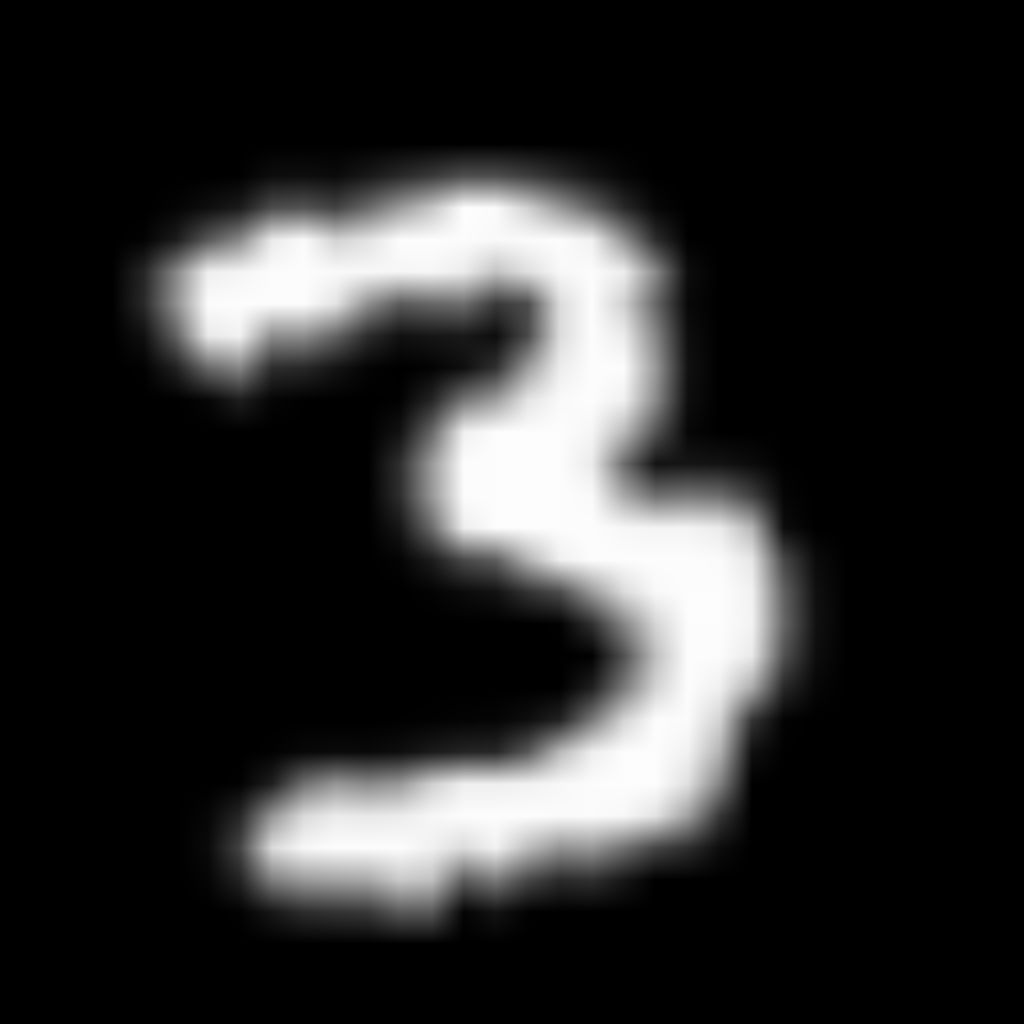}&
           
            \includegraphics[width=0.25\linewidth]{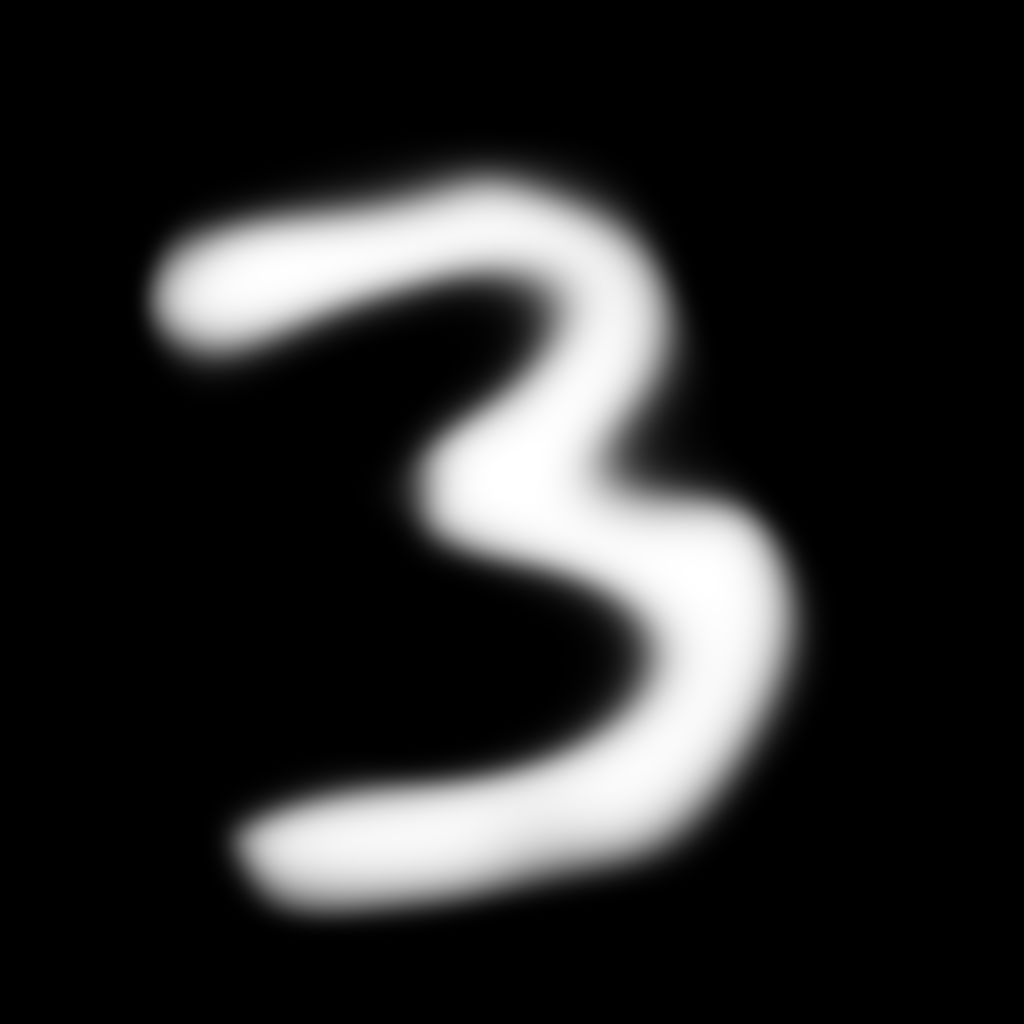}
    \end{tabular}\\
(a) Vectorization: reconstructions of MNIST images. \\
\begin{tabular}{cccc}
    Target & L1  & Perceptual  & Spiral~\cite{ganin2018synthesizing} \\
    \includegraphics[width=0.19\linewidth]{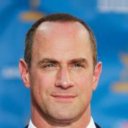}&
    \includegraphics[width=0.19\linewidth]{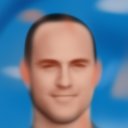}&
    \includegraphics[width=0.19\linewidth]{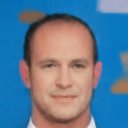}&
    \includegraphics[width=0.19\linewidth]{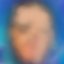}
    \\
    \includegraphics[width=0.19\linewidth]{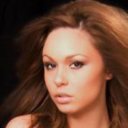}&
    \includegraphics[width=0.19\linewidth]{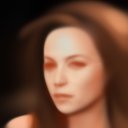}&
    \includegraphics[width=0.19\linewidth]{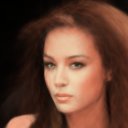}&
    \includegraphics[width=0.19\linewidth]{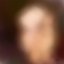}
    
    \end{tabular}
    (b) Comparison with SPIRAL~\cite{ganin2018synthesizing} on CelebA.
    \vspace{-2mm}
    \caption{Our model learns a vectorized mask representation that can be generated at any resolution without interpolation artifacts. }
    \label{fig:spiral_comp}
\end{figure}

Producing vectorized images is often essential for design applications.
We demonstrate in Fig.~\ref{fig:spiral_comp}(a) the potential of our approach for producing a continuous vector image from a low resolution bitmap. Here, we train our network on the MNIST dataset  ($28\times28$), but generate the output at resolution $1024\times1024$.
Compared to bilinear interpolation, the image we generate presents less artifacts.

We finally compare our model with SPIRAL~\cite{ganin2018synthesizing} on a few images from CelebA dataset published in \cite{ganin2018synthesizing}. SPIRAL is the approach the most closely related to ours 
in the sense that it is an iterative deep approach for reconstructing an image and extracting its structure only using a few color strokes and that it can produce vector results. We report SPIRAL results using 20-step episodes. In each episode, a tuple of 8 discrete decisions is estimated, resulting in a total of 160 parameters for reconstruction. Our results shown in Fig.~\ref{fig:spiral_comp}(b) are obtained with a model using 10  iterations and 10 mask parameters. Although we do not reproduce the stroke gesture for drawing each mask as it is the case in SPIRAL, our results reconstruct the original images much better.

\subsection{Architecture and training choices}
\label{sec:choices}

\begin{figure}[h]
    \centering
    \includegraphics[width=0.19\linewidth]{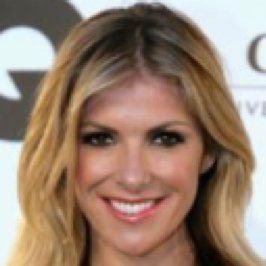}
    \includegraphics[width=0.19\linewidth]{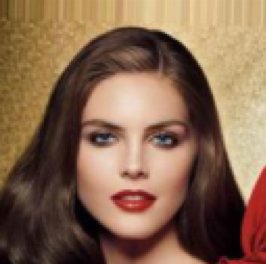}
    \includegraphics[width=0.19\linewidth]{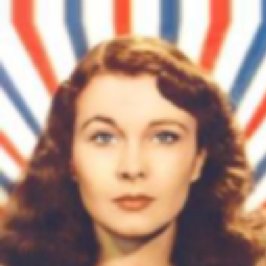}
    \includegraphics[width=0.19\linewidth]{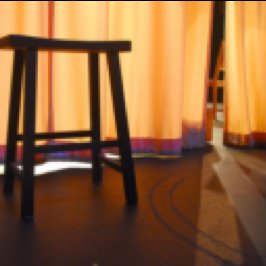}
    \includegraphics[width=0.19\linewidth]{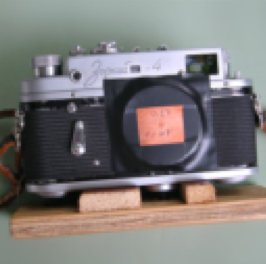}\\
    
    \includegraphics[width=0.19\linewidth]{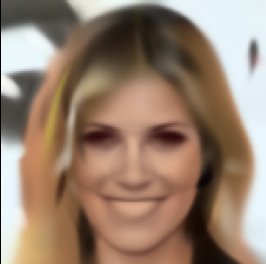}
    \includegraphics[width=0.19\linewidth]{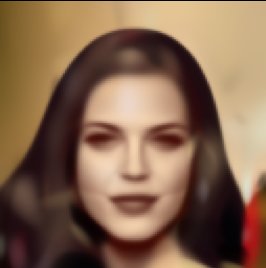}
    \includegraphics[width=0.19\linewidth]{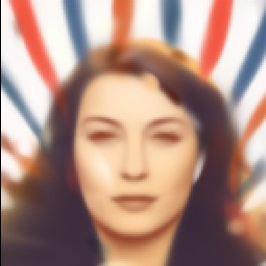}
    \includegraphics[width=0.19\linewidth]{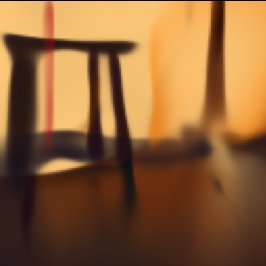}
    \includegraphics[width=0.19\linewidth]{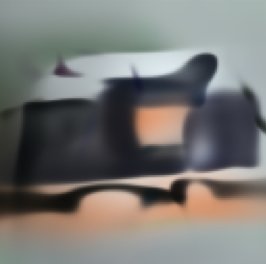}\\
    
    \includegraphics[width=0.19\linewidth]{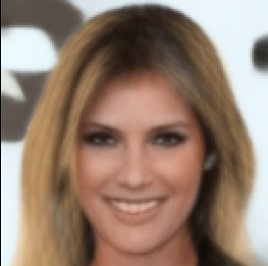}
    \includegraphics[width=0.19\linewidth]{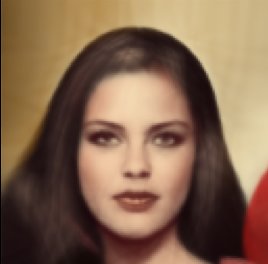}
    \includegraphics[width=0.19\linewidth]{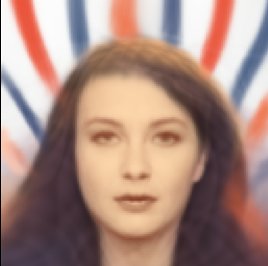}
    \includegraphics[width=0.19\linewidth]{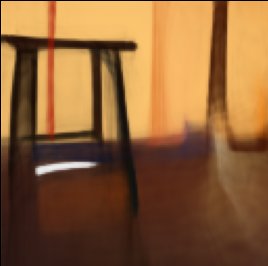}
    \includegraphics[width=0.19\linewidth]{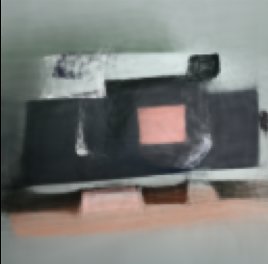}\\
    
    \includegraphics[width=0.19\linewidth]{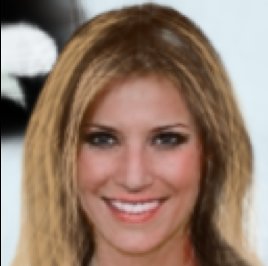}
    \includegraphics[width=0.19\linewidth]{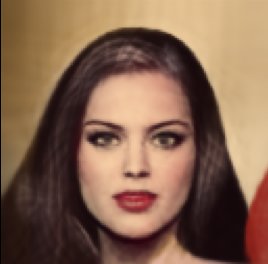}
    \includegraphics[width=0.19\linewidth]{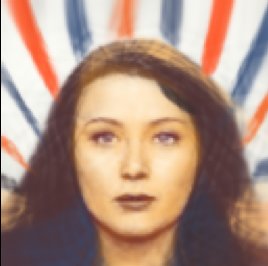}
    \includegraphics[width=0.19\linewidth]{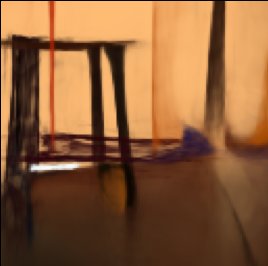}
    \includegraphics[width=0.19\linewidth]{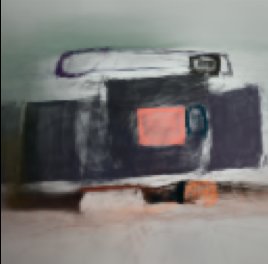}\\

    \caption{Training with perceptual and adversarial loss allows our model to reach more convincing details in the reconstructions. From top to bottom: Original images, $\ell_1$ reconstruction, using perceptual loss, adding adversarial loss.}
    
    \label{fig:L1_vs_perceptual}
\end{figure}

\paragraph{L1, perceptual and adversarial loss.}
In Fig.~\ref{fig:L1_vs_perceptual}, we show how the perceptual loss allows to obtain qualitatively better reconstructions than these obtained with an $\ell_1$ loss. Training our model with an additional adversarial loss enhances further the sharpness of the reconstructions.

In the remainder of this section, we trained our models with an $\ell_1$ loss which results in easier quality assessment using standard image similarity metrics.

\paragraph{Comparison to baselines.} 

\begin{figure}
\begin{center}

      \begin{minipage}[l]{.25\linewidth}A.\\ Resnet AE \end{minipage}  \begin{minipage}[l]{.73\linewidth}\includegraphics[width=\linewidth]{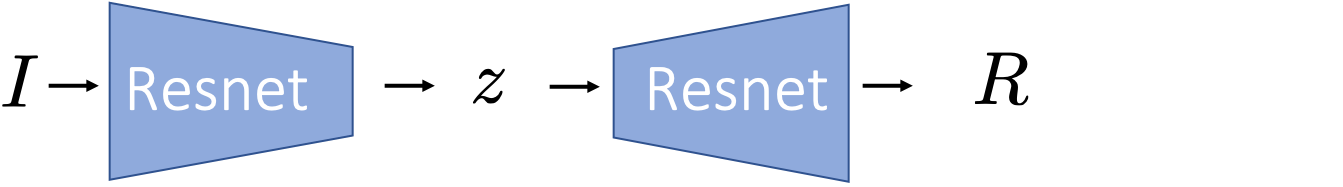}\end{minipage}\\[1mm]

   \begin{minipage}[l]{.25\linewidth}
     B.\\ MLP AE 
     \end{minipage}
     \begin{minipage}[l]{.73\linewidth}
     \includegraphics[width=\linewidth]{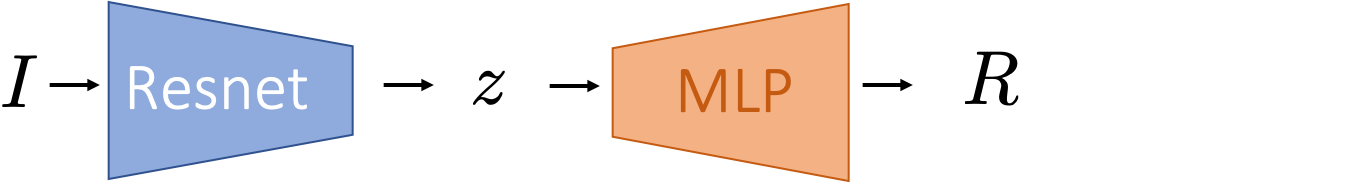}
     \end{minipage}\\[1mm]

     \begin{minipage}[l]{.25\linewidth}C. \\ Vect. AE\end{minipage} \begin{minipage}[l]{.73\linewidth}\includegraphics[width=\linewidth]{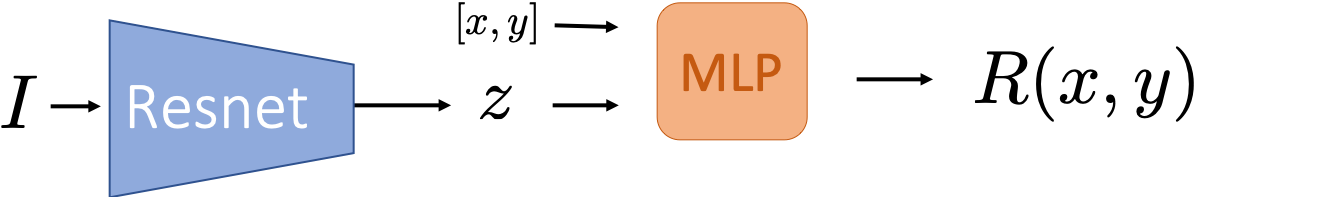}\end{minipage}\\[1mm]
     
     \begin{minipage}[l]{.25\linewidth}D. Ours\\ Oneshot\end{minipage} 
     \begin{minipage}[l]{.73\linewidth} 
     \includegraphics[width=\linewidth]{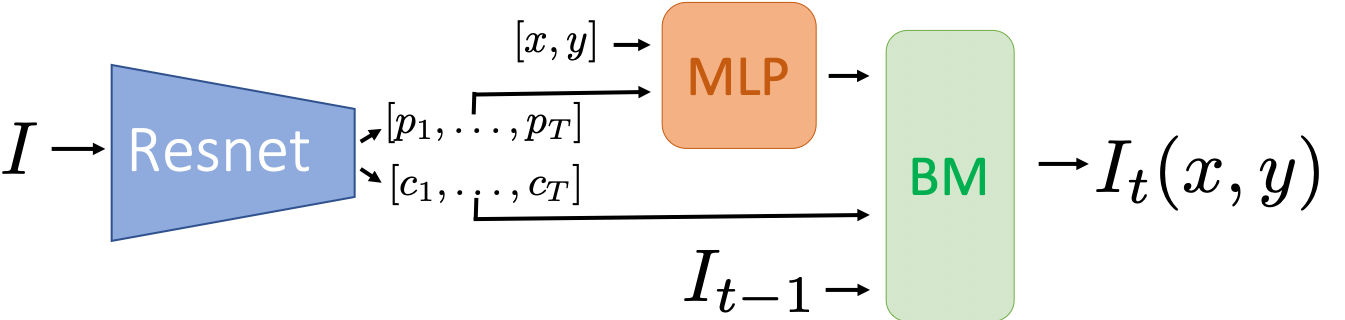}\end{minipage}\\[1mm]

     \begin{minipage}[l]{.25\linewidth}E. Ours \\ Resnet \end{minipage}
     \begin{minipage}[l]{.73\linewidth} 
     \includegraphics[width=\linewidth]{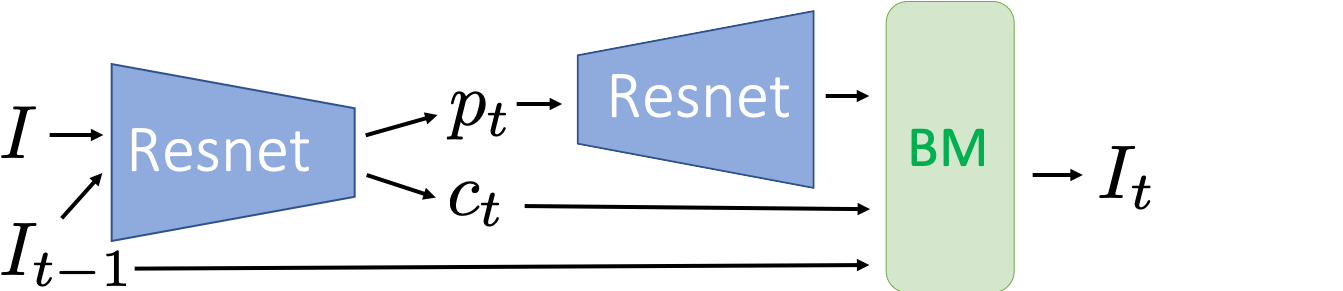}\end{minipage}\\[1mm]
     
      \begin{minipage}[l]{.25\linewidth}F. Ours \end{minipage}
      \begin{minipage}[l]{.73\linewidth} \includegraphics[width=\linewidth]{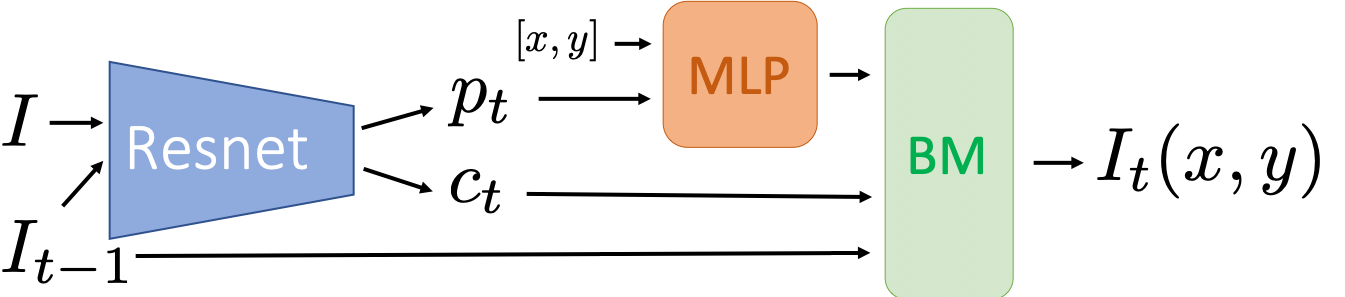}\end{minipage}\\[3mm]

    \begin{tabular}{lccc}
    \hline
                       & Layered     & Recursive       & Vectorized \\ \hline
        A. Resnet AE     &   -         &     -       &    -           \\
        B. MLP AE    &   -         &     -       &    -           \\  
        C. Vect. AE      &   -         &     -       &    \cmark      \\ 
        D. Ours Oneshot      &  \cmark     &    -        &    \cmark      \\
        E. Ours Resnet  &  \cmark     &   \cmark    &    -           \\ 
        F. Ours         &  \cmark     &   \cmark    &    \cmark      \\ \hline
    \end{tabular}
    
    \end{center}
    \vspace{-2ex}
     \caption{Considered baselines and their properties. First three baselines are non iterative. Next two are ablations of each property we consider in our final model F. Legend: $R$: reconstructed image; BM: Mask blending module of Fig. \ref{fig:mask_generation_and_blending}.}
     \label{tab:baselines_properties}
\end{figure}
As discussed in Section \ref{ssec:discussion}, every component of our model is important to obtain reconstructions similar to the target. To show that,
we provide comparisons between different versions of our model and baselines using PSNR and MS-SSIM metrics. Each baseline consists of an auto-encoder where the encoder is a residual network (ResNet-18, same as our model) producing a latent code $z$ and different types of decoders. The different baselines, depicted in Fig.~\ref{tab:baselines_properties} with a summary of their properties, are designed to validate each component of our architecture:
\begin{enumerate}[label=\Alph*.]
    \item {\it ResNet AE:} using as a decoder a ResNet with convolutions, residual connections, and upsampling similarly to the architecture used in~\cite{miyato2018spectral,kurach2018gan}.
    \item {\it MLP AE:} using as decoder an MLP with a $3\times W\times H$ output. 
    \item {\it Vect. AE:} the decoder computes the resulting image $R$ as a function $f$ of the coordinates $(x,y)$ of a pixel in image space and the latent code $z$ as $R(x,y) = f(x,y,z)$. Here $f$ is an MLP similar to the one used in our mask generation network, but with a 3-channel output instead of a 1-channel as for the mask.
    \item {\it Ours One-shot:} generates all the mask parameters $\mathbf{p_t}$ and colors $\mathbf{c}_t$ in one pass, instead of recursively. The MLP then processes each $\mathbf{p_t}$ separately leading to different masks to be assembled in the blending module as in our approach.
    \item {\it Ours ResNet:} using a ResNet decoder to generate masks $M_t$ and otherwise similar to our method, iteratively blending the masks with one color onto the canvas, in our experiments we started with a black canvas.
\end{enumerate}

Table~\ref{tab:comp_baselines} shows a quantitative comparison of results obtained by our model and baselines trained with $\ell_1$ loss and for the same bottleneck $|z|=320$. This corresponds to a size of parameters of $P=|z|/N - 3$ where 3 is the number of parameters used for color prediction.
On both datasets, our approach (F) clearly outperforms the baselines which produce vector outputs, either in one layer (C) or with one-shot parameters prediction (D). Interestingly, a parametric generation (C) is itself better than directly using an MLP to predict pixel values (B). Finally, our approach (F) has quantitative reconstruction results similar to the ResNet baselines (A and E). 
 
\begin{figure}[htb]
    \centering
    \setlength{\tabcolsep}{1pt}
    \begin{tabular}{cccccc}
    
    \includegraphics[width=0.15\linewidth]{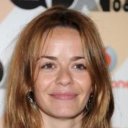}&
    \includegraphics[width=0.15\linewidth]{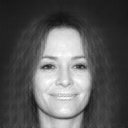}&
    \includegraphics[width=0.15\linewidth]{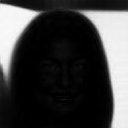}&
    \includegraphics[width=0.15\linewidth]{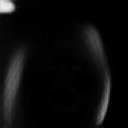}&
    \includegraphics[width=0.15\linewidth]{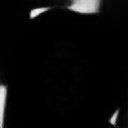}&
    \includegraphics[width=0.15\linewidth]{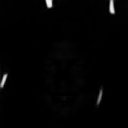}\\
    
    &
    \includegraphics[width=0.15\linewidth]{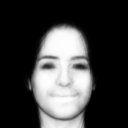}&
    \includegraphics[width=0.15\linewidth]{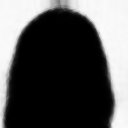}&
    \includegraphics[width=0.15\linewidth]{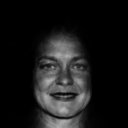}&
    \includegraphics[width=0.15\linewidth]{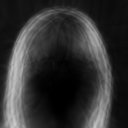}&
    \includegraphics[width=0.15\linewidth]{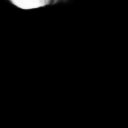}\\
    
    \end{tabular}
    \vspace{-2mm}
    \caption{Comparison of masks obtained with our approach (F) (bottom), with these obtained by our iterative ResNet baseline (E) (top).}
    \label{fig:iter_resnet_masks}
\end{figure}

However, our method has two strong advantages over Resnet generations. First, it produces vector outputs. Second, it produces more interpretable masks.  
This can be seen in Fig.~\ref{fig:iter_resnet_masks} where we compare the masks resulting from (E) and (F). Our method (F) captures much better the different components of face images, notably the hairs, while the masks of (E) include several different component in the image, with a first mask covering both hairs and faces.
In the supplementary material, we show that a qualitatively similar difference can be observed for our method when reducing the number of masks while increasing their number of parameters to keep a constant total code size.

\begin{table}[htb]
    \centering
    
    \begin{tabular}{lcccc}
    \hline
    & \multicolumn{2}{c}{ImageNet} &  \multicolumn{2}{c}{CelebA} \\
            & PSNR & MSSIM & PSNR & MSSIM \\ \hline
    
    A. MLP AE  &  16.45 & 0.46 & 19.69 & 0.78 \\
    C. Vect. AE  & 17.95 & 0.62 & 20.99 & 0.82 \\  
    D. Ours One-shot  &  20.00 & 0.77 & 23.13 & 0.89 \\
    E. Ours Resnet  & \bf 21.05 & \bf 0.82 & \bf24.67 & \bf 0.92 \\
    F. Ours   &  \bf 21.03 & \bf 0.82 & 24.02 & 0.90 \\ 
\hline
    \end{tabular}
    
    \caption{Comparing the quality of reconstruction on ImageNet and CelebA using a bottleneck $z$ of size 320 (10 masks for iterative approaches).}
    
    \label{tab:comp_baselines}
\end{table}

\begin{table}[htb]
    \centering
    \setlength\tabcolsep{3pt}
    \vspace{-1mm}
    \begin{tabular}{|p{1.2cm}|c|c|c|c|c|c|}
    \hline
    & \multicolumn{2}{c|}{$T=5$} &  \multicolumn{2}{c|}{$T=10$} &  \multicolumn{2}{c|}{$T=20$} \\ \hline
      & \small{One-shot}  & \small{Ours}   & \small{One-shot} & \small{Ours}    & \small{One-shot}   & \small{Ours} \\ \hline
    \small{PSNR}  & 21.97 & 23.07   & 22.25 & 24.2  & 22.37 & 24 \\ \hline
    \small{Time(h)} \scriptsize{$95\%$ PSNR}  
             & 7.6 & 9.8  & 12.1 & 16.9 & 19.8 & 36.5  \\ \hline
    \small{Testing time} \scriptsize{(ms)}  & 12 &  32  & 18 &  65 & 31 & 129  \\
             \hline
    \end{tabular}
    \caption{Comparison of our recursive strategy with the One-shot approach, in terms of reconstruction quality (PSNR) and training time required to reach $95\%$ of its best achievable PSNR at full convergence on CelebA. The inference time does not exceeds 0.2 seconds.}
     \label{tab:recursive_vs_oneshot}
\end{table}

\paragraph{Recursive setup and computational cost.}
There is of course a computational cost to our recursive approach. In Table \ref{tab:recursive_vs_oneshot}, we compare the PSNR and computation time for the same total number of parameters (320) but using different number of masks $T$, both for our approach and the one-shot baseline. Interestingly, the quality of the reconstruction improves with the number of masks for both approaches, our approach being consistently more than a point PSNR better than the one-shot baseline. However, as expected our approach is slower than the baseline both for training and testing, and the cost increases with the number of masks. 

Table \ref{tab:more_forward} evaluates the reconstruction quality when recomposing images at test time with a larger number of masks $N$ than the $T=10$ masks used at training time. On both datasets, the PSNR increases by almost a point with additional masks. This is another advantage of our recursive approach.

\begin{table}[h]
    \centering
    \setlength\tabcolsep{2.4pt}
    \resizebox{\linewidth}{!}{
    \begin{tabular}{|p{0.9cm}|c|c|c|c|c|c|c|c|}
    \hline
       &   \multicolumn{4}{c|}{ImageNet}  & \multicolumn{4}{c|}{CelebA}\\ \hline
    \small{$N$}   &    10 & 20 & 40 & 80 & 10 & 20 & 40 & 80 \\ \hline
    \small{PSNR}   &    20.97 & 21.72 & 21.83 & 21.84 & 24.02 & 24.82 & 24.86 & 24.86 \\
    \hline
    \end{tabular}
    }
    \caption{Forwarding more masks at test time improves reconstruction. These results with $N$ masks forwards are obtained with a model trained using $T$=10 masks ($|z|=320$).}
    \label{tab:more_forward}
\end{table}

\section{Conclusion}

We have presented a new paradigm for image reconstruction using a succession of single-color parametric layers. This decomposition, learned without supervision, enables image editing from the masks generated for each layer. 

We also show how the learned mask parameters may be exploited in a retrieval context. 
Moreover, our experiments prove that our image reconstruction results are competitive with convolution-based auto-encoders.

Our work is the first to showcase the potential of a deep vector image architecture for real world applications. 
Furthermore, while our model is introduced in an image reconstruction setting, it may be extended to adversarial image generation, where generating high resolution images is challenging. We're aware of risks surrounding manipulated media but we believe the importance of publishing this work openly may have benefits in AR filters or more realistic VR.

We think that because of its differences and its advantages for user interaction, our method will inspire new approaches.

{\small
\bibliographystyle{ieee}
\bibliography{main}
}

\end{document}